# Phase-Adaptive LLM Framework with Multi-Stage Validation for Construction Robot Task Allocation: A Systematic Benchmark Against Traditional Optimization Algorithms

Shyam prasad reddy Kaitha[1] and Hongrui Yu[2]

[1] PhD Student, Department of Civil and Environmental Engineering, Virginia Tech Email: skaitha@vt.edu

[2] Assistant Professor, Department of Civil and Environmental Engineering, Virginia Tech Email: hryu42@vt.edu

## ABSTRACT

Multi-robot task allocation in construction automation has traditionally relied on optimization algorithms such as Dynamic Programming and Reinforcement Learning methods. This research presents the LangGraph-based Task Allocation Agent (LTAA) framework, introducing phase-adaptive allocation strategies, multi-stage validation with hierarchical retry mechanisms, and dynamic prompting for efficient construction robot coordination. While recent Large Language Models (LLM) approaches show promise for construction robotics coordination, they lack rigorous validation and performance benchmarking against established methods. This paper presents the first systematic comparison of LLM-driven task allocation against traditional algorithmic approaches in construction scenarios. Through systematic framework development validating LLM feasibility via SMART-LLM replication and addressing implementation challenges through Self Corrective-Agent Architecture. The authors developed the LangGraph-based Task Allocation Agent (LTAA) framework, an LLM-driven coordination system that combines natural language reasoning with phase-adaptive allocation strategies and hierarchical validation mechanisms. The framework also offers major computational efficiencies, reducing token usage by 94.6% and allocation time by 86% through dynamic prompting.The framework adapts its allocation strategy across phases: prioritizing execution feasibility in initial assignments, then emphasizing workload balance in subsequent allocations.The authors evaluate LTAA framework against Dynamic Programming, Q-learning, and Deep Q-Network (DQN) baselines using construction operations from a benchmark human-robot collaboration dataset - TEACh, across various task allocation tasks. LTAA framework achieves 76% task completion rate, performing competitively with Q-learning (73%) and DQN (77%). In the Heavy Excels configuration where robots have pronounced specializations ,LTAA framework reaches 77% completion with superior workload balance, exceeding all traditional methods. These results demonstrate that LLM-based reasoning with structured validation can match optimization algorithms for construction task allocation, establishing LLM-driven approaches as viable alternatives that offer additional benefits of natural language interpretability and rapid adaptability to changing requirements without retraining. Future work will explore additional phase-adaptive trade-off strategies and investigate domain-specific fine-tuning of LLMs to improve allocation reasoning for construction-specific constraints.

## INTRODUCTION

The construction industry faces substantial challenges in workforce capacity, productivity growth, and worker safety. The U.S. construction industry requires an estimated 501,000 additional workers beyond normal hiring to meet 2024 demand (Associated Builders and Contractors 2024). These workforce shortages directly result in project delays and increased costs (Delvinne et al. 2020, Sokas et al. 2019). Furthermore, construction accidents significantly impact worker well-being and project outcomes (Fontaneda et al. 2022). Construction productivity has grown at only 1% annually compared to 2.8% for the total economy (McKinsey Global Institute 2024). Moreover, the industry accounts for over 1,000 workplace fatalities annually (Bureau of Labor Statistics 2023). These quantified challenges necessitate innovative approaches to maintain construction capability and meet growing infrastructure demands.

Construction robotics presents significant potential to address workforce gaps while enhancing safety and productivity outcomes. Robots can play a crucial role in addressing workforce gaps and mitigating challenges posed by labor shortages (Lundeen et al. 2018, Wang et al. 2021, Brosque , Fischer 2022, Park et al. 2023 and Yu et al. 2023). Furthermore, robots possess superior physical capabilities and excel in handling heavy and repetitive construction tasks while being less prone to physical fatigue and cognitive lapses (Liang et al. 2021, 2023). As a result, deploying construction robots achieves significant improvements in construction productivity, leading to reduced delays and construction costs (Pan and Pan 2020, Ryu et al. 2021, Liu et al. 2024 and Chandramouli et al. 2024). However, individual robots cannot complete all tasks due to the complexity of construction work (Pan et al. 2020, Ye et al. 2024, Yu et al. 2025 and Fu et al. 2022). Complete construction workflows require diverse capabilities that exceed any individual robot's specialized skill set. For instance, a wall construction task may require material transport, precise positioning, welding, and quality inspection. Consequently, single-robot execution of complete workflows is impractical and necessitates coordinated multi-robot collaboration.

Multi-Robot Task Allocation (MRTA) decomposes complicated tasks into reasonable subtasks and assigns them to robots. This process considers constraints such as robot capabilities, task requirements, and environmental conditions to achieve optimal matching (Dai et al. 2020, Ye et al. 2024). MRTA problems are characterized along dimensions of single-task versus multi-task robots, single-robot versus multi-robot tasks, and instantaneous versus time-extended assignments (Gerkey and Matarić 2004, Korsah et al. 2013). The construction coordination challenges addressed in this research fall into the most complex category: multi-robot, multi-task, time-extended allocation. These challenges require sophisticated reasoning approaches that can handle heterogeneous robot capabilities, diverse task requirements, and temporal dependencies across project phases. However, the MRTA problem is mainly studied in warehouse logistics or environmental exploration and rarely addressed specifically for construction industry applications (Ye et al. 2024, Dai et al. 2020).

Traditional MRTA approaches have achieved success across various domains through optimization algorithms. These include linear programming methods, genetic algorithms, and Re-inforcement Learning (RL) that provide mathematical rigor and proven performance guarantees (Kuhn 1955, Atay 2006, Jones 2020, Chen 2019). However, construction environments present fundamentally different challenges that render traditional optimization approaches insufficient, as the unstructured construction sites create inherent coordination difficulties (Feng et al. 2016, Liang and Cheng 2023). Moreover, the complexity and information abundance in construction contexts exceed the reasoning capabilities of mathematical optimization methods (Makondo et al. 2015, Xu et al. 2020). This inadequacy of traditional approaches for construction coordination has created the need for more adaptive solutions.

Recent advances in Large Language Models (LLM) demonstrate remarkable capabilities in natural language understanding, logical reasoning, and adaptive decision-making. These capabilities address several fundamental limitations of traditional optimization and learning-based approaches (Kannan et al. 2024). Furthermore, LLM-based coordination approaches enable intuitive human-robot interaction and adaptive coordination strategies capable of handling complex, dynamic requirements characteristic of construction environments. Recent LLM frameworks have shown promise through task planning, digital twin integration, and multi-agent architectures (Kannan et al. 2024, Prieto et al. 2024, Deng et al. 2025). However, existing frameworks lack systematic validation mechanisms and performance benchmarking against traditional methods. Consequently, a critical gap exists: while LLM-based coordination shows promise, no research has developed iterative retry and self-corrective mechanisms to ensure reasoning reliability, nor subsequently conducted systematic benchmarking to compare LLM performance against traditional optimization methods for construction task allocation.

This study addresses this research gap by proposing the LangGraph based Task Allocation Agent (LTAA) framework as a novel LLM-driven coordination approach introducing novel phase-adaptive strategies and multi-stage validation mechanisms absent in existing frameworks. The phase-adaptive strategy demonstrates effectiveness in achieving workload balance without significant performance sacrifice, while the multi-stage validation system maintains reasoning consistency by systematically detecting and correcting allocation deficiencies through structured feedback loops, for construction robotics. To demonstrate its effectiveness and validate its performance, the authors systematically evaluate the framework against established algorithmic and learning-based baselines through comprehensive benchmarking. Building upon established probabilistic robot modeling principles and construction-specific constraints, this research provides the first systematic performance comparison between LLM-based reasoning and traditional approaches for construction robotics task allocation. The LTAA framework incorporates capability-aware modeling, phase-adaptive allocation strategies, and validated LLM decision-making to address construction-specific coordination challenges identified in prior work (Bock 2015, Delgado et al. 2019). Through systematic comparison against traditional algorithmic methods (Brute Force, Greedy Method) and RL approaches (Q-learning, Deep Q-Network (DQN)), this study contributes both a methodologically rigorous LLM coordination system and empirical insights into the comparative effectiveness of reasoning-based versus traditional approaches for construction automation.

The rest of this paper is structured as follows. Section 2 reviews multi-robot task allocation literature across optimization, learning-based, and LLM-driven approaches. Section 3 presents the LTAA framework methodology. Section 4 describes the experimental setup. Section 5 presents comparative results against traditional algorithmic methods (Brute Force, Greedy Method) and RL approaches (Q-learning, DQN). Section 6 concludes with findings and future directions.

## LITERATURE REVIEW

### Multi-Robot Task Allocation (MRTA)

MRTA provides the theoretical foundation for coordinating robotic teams through systematic task distribution considering resource constraints, temporal requirements, and robot capability heterogeneity. Gerkey and Matarić (2004) established the foundational MRTA taxonomy characterizing allocation problems along three critical dimensions: single-task versus multi-task robots, single-robot versus multi-robot tasks, and instantaneous versus time-extended assignments. This taxonomy enabled systematic analysis of coordination complexity and algorithmic requirements across different problem formulations. However, the original taxonomy proved insufficient for addressing heterogeneous robot capabilities and temporal dependencies characteristic of real-world applications. Consequently, (Korsah et al. 2013) developed extended frameworks incorporating robot capability diversity, task interdependencies, and dynamic constraint satisfaction. This extended framework reflected the field's progression toward heterogeneous teams executing complex, interconnected workflows.

While these enhanced taxonomic frameworks acknowledged that practical deployment scenarios involve robots with specialized capabilities executing complex tasks, construction environments present unique MRTA challenges. These challenges distinguish construction from controlled settings where traditional approaches achieved success. Construction sites involve unstructured layouts, dynamic material flows, weather dependencies, and complex interdependencies between trades (Yu et al. 2013). Moreover, these environments demand allocation frameworks capable of handling spatial constraints, temporal dependencies, and safety-critical operations (Garcia de Soto et al. 2023, Chakraa et al. 2023). Furthermore, construction requires human oversight mechanisms that cannot be predetermined or easily encoded in mathematical formulations. These construction-specific challenges necessitate adaptive coordination approaches capable of handling dynamic constraints and unstructured environments. Traditional optimization methods have attempted to address these MRTA challenges through various algorithmic strategies.

**Traditional Optimization Approaches for MRTA**

Classical optimization methods form the backbone of established multi-robot coordination approaches, providing mathematical rigor and performance guarantees through various computational techniques. Kuhn (1955) established the theoretical basis for optimal assignment problems through the seminal Hungarian method. This method demonstrated that bipartite matching between tasks and resources could be solved efficiently using linear programming with polynomial time complexity guarantees. However, the Hungarian method was limited to simple one-to-one assignments and could not handle heterogeneous robot capabilities or task dependencies. In construction environments, robots possess specialized capabilities such as heavy lifting versus precision assembly, and workflows require coordinated task sequences with explicit dependencies. Consequently, these limitations motivated more sophisticated optimization formulations.

Atay and Bayazit (2006) pioneered mixed-integer linear programming (MILP) formulations for multi-robot task allocation, incorporating robot capability constraints, task precedence relationships, and communication limitations. Military and aerospace applications further advanced MILP-based coordination through sophisticated formulations addressing heterogeneous capabilities and timing constraints (Darrah et al. 2005, Schumacher et al. 2004). MILP provided significant advantages in modeling complex multi-robot scenarios with explicit constraint satisfaction. However, MILP approaches rely on centralized optimization architectures that face scalability challenges when coordinating large robot teams. Construction projects involve multiple robots distributed across large, unstructured sites with limited communication infrastructure, making centralized coordination impractical. These scalability limitations motivated distributed coordination approaches.

Advanced market mechanisms demonstrated significant advantages in scalability, adaptability, and fault tolerance compared to centralized approaches. Botelho and Alami (1999) pioneered structured auction protocols enabling competitive task assignment, while Gerkey and Matarić (2002) demonstrated that auction methods could achieve effective task distribution while preserving coordination properties. Recent advances have addressed communication constraints through sophisticated bidding mechanisms under limited connectivity conditions (Ferri et al. 2017, Quinton et al. 2022). However, auction-based methods may not guarantee global optimality, as local bidding decisions can lead to suboptimal coordination when tasks have complex interdependencies. Construction workflows require precise coordination between multiple trades where task sequences must follow specific orders, and auction mechanisms struggle with such temporal constraints critical for project scheduling. These coordination challenges motivated exploration of structured optimization approaches.

Dynamic programming approaches offered alternative optimization strategies suited to sequential decision-making scenarios. Bellman (1962) established the theoretical framework providing optimal solution methods for problems exhibiting optimal substructure properties. However, dynamic programming faces exponential state space growth limiting practical applicability to small problem instances. Construction projects involve numerous robots executing dozens of tasks with complex state dependencies, making computational requirements prohibitive for real-time decision-making. Consequently, researchers explored metaheuristic approaches handling large, complex solution spaces.

Population-based metaheuristic approaches leveraged evolutionary principles to navigate complex solution spaces. Genetic algorithms demonstrated effectiveness for handling combinatorial complexity and constraint hierarchies without requiring complete mathematical problem formulation (Jones et al. 2010, Al-Omeer and Ahmed 2019). Swarm intelligence methods provided distributed optimization capabilities suited to coordination scenarios involving large robot teams (Chen et al. 2022, Lim and Isa 2015, Wang et al. 2012, Blum 2005). However, metaheuristic approaches provide no guarantees of solution optimality and remain computationally expensive, requiring numerous iterations to achieve acceptable solution quality. In construction environments requiring real-time adaptation to changing conditions, the computational time for iterative optimization becomes impractical.

Despite the evolution of these optimization approaches, all traditional methods share fundamental limitations when applied to construction coordination scenarios. The requirement for complete problem specification at planning time conflicts with construction's inherently dynamic nature, where task requirements, environmental conditions, and resource availability change continuously throughout project execution. Furthermore, computational complexity often becomes prohibitive for large-scale construction projects involving numerous robots and complex task dependencies. These persistent limitations motivated researchers to explore learning-based coordination approaches that could adapt to dynamic environments without requiring complete problem specification.

**Learning-Based and AI-Driven Coordination Approaches**

RL has emerged as a transformative approach to multi-robot task allocation through its capability to learn optimal coordination policies from environmental interaction without requiring complete problem specification. RL frameworks enable robots to develop coordination expertise through trial-and-error learning while adapting to environmental changes and evolving task requirements (Chen et al. 2019, Arulkumaran et al. 2017). However, RL requires extensive training through repeated environmental interaction and trial-and-error exploration. In construction environments, this exploration process poses significant safety risks where coordination errors could endanger workers or damage expensive equipment. Furthermore, the time required for training coordination policies is impractical for construction projects with tight schedules. These safety and efficiency concerns motivated construction-specific RL approaches.

Lee et al. 2022 developed a digital twin-driven Deep Reinforcement Learning (DRL) approach specifically for adaptive task allocation in robotic construction. This approach integrated Building Information Modeling (BIM) with DRL to enable context-aware coordination decisions incorporating spatial constraints, temporal dependencies, and safety requirements. The digital twin framework enabled safer policy learning in simulation before real-world deployment. However, the exploration requirements inherent in RL still conflict with construction safety requirements where even simulation-trained policies may produce dangerous coordination failures during real-world deployment (Zhao et al. 2020, Li 2017). Moreover, transferring learned policies from simulation to real construction sites with different environmental conditions remains challenging. These sim-to-real transfer challenges motivated advanced RL techniques

Advanced RL techniques incorporated domain randomization and reward shaping to address simulation-reality gaps. (Tobin et al. 2017) demonstrated that domain randomization could enable successful sim-to-real transfer for robotic tasks, while (Grzes and Kudenko 2010) developed reward shaping approaches enhancing learning efficiency in multi-agent scenarios. These techniques improved robustness to environmental variability in construction robotics. However, these approaches still require extensive computational resources and careful hyperparameter tuning. In dynamic construction environments where conditions change frequently, the computational overhead and tuning complexity become impractical for real-time coordination. Furthermore, learned policies remain difficult to interpret and validate. These interpretability limitations motivated game-theoretic approaches.

Game-theoretic approaches provided sophisticated mathematical frameworks for modeling strategic interactions among autonomous agents. Martin et al. 2023 developed multi-robot task allocation clustering based on game theory, employing Shapley values to measure individual robot contributions. Their framework demonstrated superior performance while providing formal analysis of equilibrium solutions and stability properties. However, game-theoretic approaches still lack interpretability in decision-making processes and assume rational agent behavior that may not hold in uncertain construction environments. Moreover, they require precise utility function specifications that are difficult to define for complex construction tasks.

Despite these advances, all learning-based and AI-driven approaches face critical limitations when applied to construction coordination scenarios requiring explainable decision-making. The black-box nature of deep learning and game-theoretic models prevents clear understanding of coordination decision processes (You et al. 2023). Construction environments require human supervisors to understand and validate coordination rationales for safety oversight and project management. These explainability limitations motivated exploration of Large Language Model-based coordination approaches capable of providing natural language reasoning and transparent decision-making processes.

**Large Language Model-Based Coordination Paradigms**

LLM represent a transformative paradigm for multi-robot coordination through their capabilities in natural language understanding, contextual reasoning, and adaptive decision-making. Unlike optimization methods requiring complete problem specification, LLM-based approaches enable intuitive human-robot interaction and address the explainability limitations of learning-based methods by providing transparent, natural language reasoning.

Kannan et al. 2024 developed the SMART-LLM framework for LLM-driven multi-robot task planning through their SMART-LLM framework, demonstrating that language models could effectively perform task decomposition and allocation using programmatic prompts. The framework achieved 70% success rates while maintaining interpretability through natural language reasoning. However, SMART-LLM was evaluated in household robotics with static task requirements.

Construction environments present fundamentally different challenges including dynamic site conditions, unpredictable disruptions such as material delays and weather changes, and complex task dependencies requiring real-time adaptation. These limitations motivated construction-specific LLM applications coordination scenarios while maintaining interpretability through natural language reasoning.

For construction robotics, Deng et al. 2025 developed an integrated framework combining digital twins, optimization backends, and LLM-driven narrative interpretation for dynamic construction environments. This framework addressed adaptive task rescheduling in response to material delays, site conditions, and weather disruptions, achieving over 97% accuracy in constraint extraction. However, this framework is more focused on task rescheduling after initial allocation than initial allocation optimization. These complementary requirements motivated LLM frameworks for initial construction task planning between physical construction sites and digital twin representations enables continuous system adaptation to evolving site conditions. Parallel developments have explored LLM applications for construction robot control code generation, with hierarchical generation approaches demonstrating substantial reductions in programming errors through customized API libraries and chain-of-action prompting techniques.

Multi-agent LLM architectures emerged as sophisticated approaches to construction task allocation challenges. Prieto et al. 2024 introduced collaborative frameworks employing Planner and Supervisor agents demonstrating improved reliability, while Kim et al. 2025 developed frameworks integrating BIM-based knowledge with natural language dialogue for construction applications. Despite these advances, all LLM-based coordination approaches lack systematic validation and rigorous performance benchmarking against established optimization methods. No prior research provides comprehensive performance comparison between LLM-driven task allocation and traditional algorithmic approaches for construction robotics. This critical validation gap necessitates systematic evaluation to determine whether LLM-based reasoning can achieve competitive performance with proven optimization methods while providing interpretability and adaptability benefits.

# METHODOLOGY

**Problem Definition**

MRTA in construction robotics requires strategies that adapt as priorities shift from execution reliability to workload equity throughout project progression. Traditional optimization approaches such as, Dynamic Programming (DP), Q-learning, and DQN apply static objectives that cannot adjust to evolving project context. This research develops the LTAA framework to address this limitation through phase-adaptive reasoning, systematically benchmarked against traditional algorithmic approaches.

Unlike conventional approaches that rely solely on deterministic formulations such as DP for optimal subproblem solutions, RL-based allocation such as DQN and Q-learning, the proposed LTAA framework employs context-aware reasoning to balance success probability maximization with workload fairness. For example, during the early phase of task allocation, the LLM prioritizes robots with the highest success probabilities, while in later phases it adaptively redirects tasks toward underutilized robots to restore workload balance.

This LTAA framework aims to address three key objectives:

1. Demonstrate LLM feasibility for complex optimization problems.

2. Assess the computational efficiency of LLM-driven allocation using dynamic prompting and

3. Validate LLM performance against algorithmic solutions such as DQN and Q learning.

The framework accomplishes these objectives through systematic progression from task decomposition to balanced allocation optimization. Long-horizon construction tasks are first decomposed into manageable subtasks using object-centric and skill-centric decomposition strategies adapted from the SMART-LLM framework (Kannan et al. 2024). The object-centric decomposition part identifies task components based on physical objects and their required manipulations, while the skill-centric decomposition part organizes subtasks according to robot capability requirements such as precision handling, force application, or careful maneuvering. These decomposed subtasks are then assigned to heterogeneous robots through capability-aware matching that considers robot specializations and task requirements. The LTAA framework employs LangGraph orchestration with probabilistic success modeling to maximize task completion likelihood by computing robot-specific success rates for each subtask based on feature-capability alignment. However, initial testing revealed significant workload imbalance across robot teams, where pure success-probability maximization resulted in overutilization of high-capability robots and underutilization of specialized units. This observation motivated the development of phase-adaptive allocation strategies that dynamically adjust decision priorities throughout project progression, transitioning from success-rate emphasis in early phases to workload equity prioritization in later phases, thereby balancing overall mission success with fair resource utilization.

The methodology development as shown in Fig. 1 proceeded through three stages. First, SMART-LLM implementation assessed LLM feasibility for MRTA but revealed validation inconsistencies and output reliability issues. Second, the Self Corrective-Agent Architecture addressed these challenges through multi-stage validation and hierarchical retry mechanisms. Finally, the LTAA framework integrated these validation principles with phase-adaptive allocation strategies to enable systematic benchmarking against traditional optimization methods. The authors first describe the process of implementing a classic LLM-based MRTA framework: SMART-LLM (Kannan et al. 2024), and the fundamental challenges encountered during implementation. The main challenges observed include validation inconsistencies, LLM output reliability issues, and coordination limitations. These challenges motivated the development of a phase-adaptive and self-corrective Self Corrective-Agent LangGraph Architecture, using multi-stage validation and hierarchical retry mechanisms to address the identified deficiencies. These contributions formed the proposed LTAA framework. This paper also introduced a systematic performance benchmarking against traditional optimization methods.The LTAA framework evaluation employs the TEACh dataset as the standard benchmark for direct comparison with Q-learning, DQN, and DP baselines under identical task scenarios. Evaluation criteria include workload balance distribution, feature-specific performance across task types, and reasoning quality assessment to validate both allocation effectiveness and interpretability.

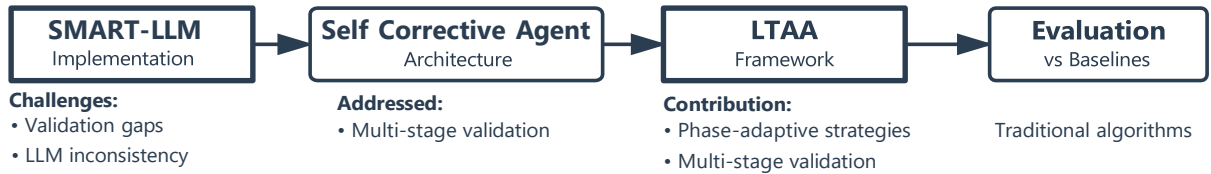

Fig. 1. Framework Development Progression

**Framework Design Rationale: SMART-LLM Implementation Study**

As mentioned, to assess the feasibility of LLM-driven reasoning for multi-robot task allocation, the authors implemented the SMART-LLM framework (Kannan et al. 2024) using the AI2-THOR simulation environment with 36 benchmark tasks. The replication process uncovered fundamental reliability challenges (detailed in Appendix 1). Of the multiple implementation challenges encountered, two most critical to framework architecture: LLM generation inconsistency and context window limitations. These challenges motivated the self-corrective approach and modular design principles embodied in LTAA framework.

**Generation Inconsistency and Capability Limitations** LLM performance varied significantly across models and execution attempts. Testing GPT-3.5, GPT-4, Calude 3.5 Haiku and Llama-70B with identical prompts revealed two critical patterns:

Two key patterns were observed:

1. Non-Deterministic Generation: Claude 3.5 Haiku and GPT-4 generated inconsistent outputs across identical prompts some executable, others containing syntax errors.

2. Training Example Contamination: Over repeated runs, GPT-4 occasionally reused few-shot example sequences, creating logical errors (e.g.,replacing "throw spatula in trash" task objects with objects of task "put eggs in fridge").

These inconsistencies revealed absent validation mechanisms in the original architecture, directly motivating explicit validation and retry systems in the LangGraph framework.

**Context Window and Token Limit Constraints**

The framework's few-shot prompting approach concatenated example demonstrations from three stages (task decomposition, allocation, and code generation) into a single unified prompt. Appending additional examples to enhance performance on complex tasks caused cumulative token counts to exceed the model's context window, triggering API failures during inference. Extended prompts for complex tasks resulted in token limit exceeded exceptions, preventing executable code generation. This scalability bottleneck where richer context improved accuracy but exceeded feasible input lengths informed the need for modular prompt optimization and dynamic context management, later integrated into the LTAA framework.

These two challenges LLM generation inconsistency and context window limitations directly shaped the architectural requirements for reliable multi-robot coordination: systematic validation mechanisms to ensure reasoning quality, and modular design to manage context constraints. To address these requirements, the authors developed a validation-centric architecture incorporating multi- stage validation, structured feedback loops, and controlled retry mechanisms.

## FRAMEWORK ARCHITECTURE

**Self Corrective-Agent LangGraph Framework Architecture**

To address the limitations identified during the SMART-LLM implementation process and to verify the practical feasibility of LLM-driven reasoning for multi-robot task allocation, a Self Corrective-Agent Framework was developed. Unlike the earlier single-agent reasoning design, this framework implements a multi-stage iterative pipeline consisting of three dedicated agents: Task Decomposition Agent, Task Allocation Agent, and Code Generation Agent. As shown in Figure 2, each agent is coupled with an independent validation node and orchestrated through the LangGraph workflow engine. The Decomposition Agent initiates the pipeline by translating high-level natural-language instructions into a structured sequence of subtasks. It identifies task dependencies, environment objects, and the temporal ordering required for successful completion. The Allocation Agent subsequently maps these validated subtasks to the most suitable robots based on their capabilities, skill sets, and available resources, while determining execution order (sequential or parallel) and coalition structure. Finally, the Code Generation Agent converts the allocation plan into fully executable Python code that integrates task sequencing, API calls, and robot motion command in the AI2THOR simulation environment. Each agent produces an output only after passing its corresponding validation node, which performs syntactic, semantic, and logical checks to ensure correctness before advancing to the next stage.

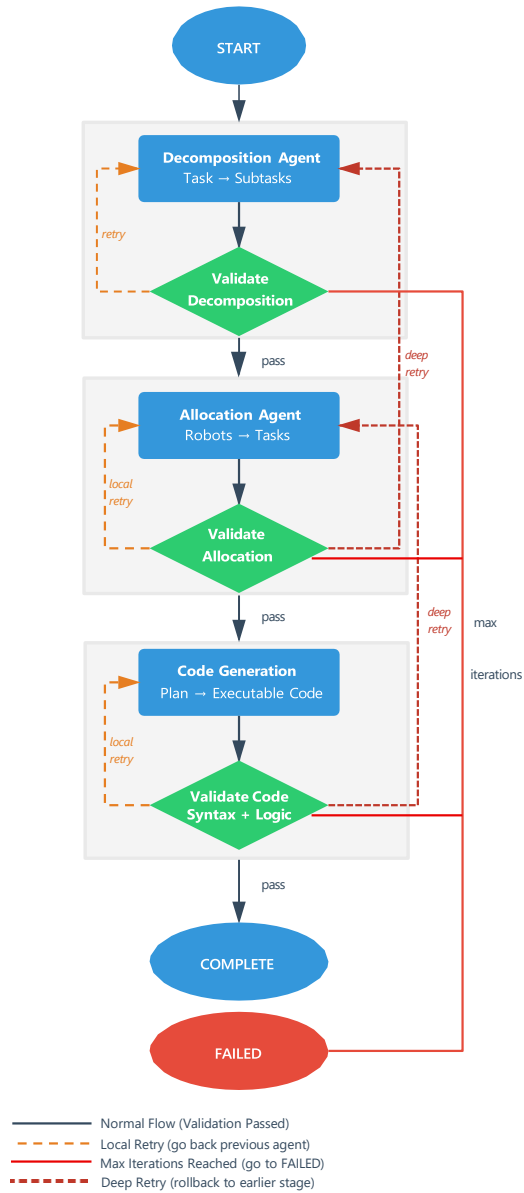

**Fig. 2.** Self Corrective-Agent Task Allocation Framework with LangGraph Orchestration

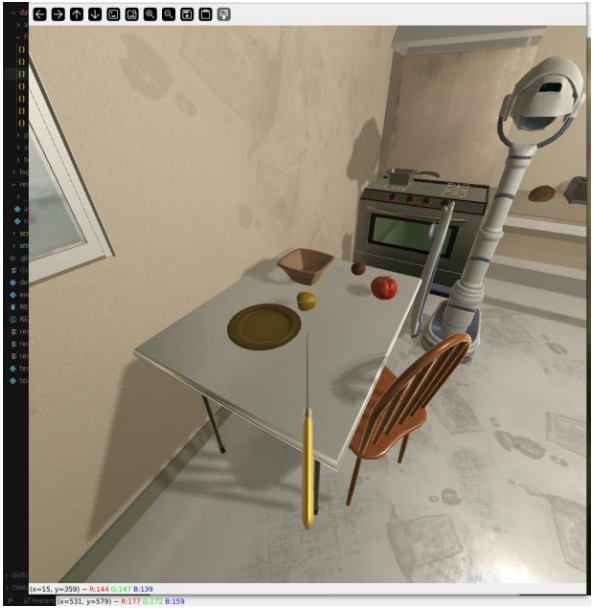

**Fig. 3.** Front view of an agent slicing an apple task

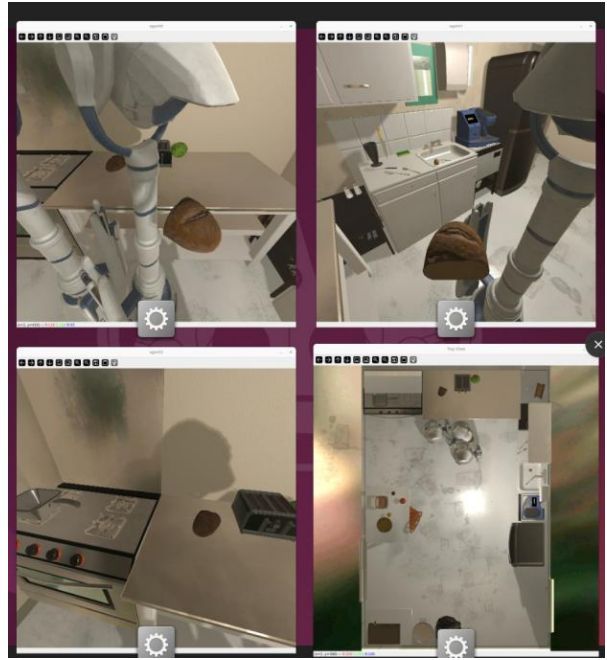

**Fig. 4.** Agent Executing Toasting Task (Multi-View)

Overall, the architecture adopts a state-driven orchestration model governed by a centralized pipeline state schema, which maintains the complete contextual history across all stages. This state management design enables continuity, feedback propagation, and deterministic termination through controlled retry logic.

The authors also introduced a systematic internal validation framework for this novel method. Validation is integrated at three levels: validate Decomposition, validate Allocation, and validate Code for robot execution ensuring that errors are detected early and localized. When a validation failure occurs, the framework triggers a structured feedback loop: 1) local retries that allow agents to regenerate outputs with corrective context, 2) deep retries that roll back to previous stages (e.g., repeated code-generation failures invoke re-allocation). Each feedback cycle generates standardized feedback messages objects containing failure reasons, contextual expectations, and suggested corrections, enabling the agents to iteratively refine their reasoning while preserving traceability.

In addition, this framework employs a multi-layer retry and iteration control mechanism to opti-mize computational resource usage. It limits each agent to a maximum of five local attempts before initiating a deep retry, and a global cap of 25 total iterations to guarantee deterministic termination. This systematic escalation strategy allows the system to recover from localized reasoning errors and to adaptively correct upstream planning faults rather than redundancy regeneration. Moreover, a Routing Logic component was introduced to govern the progression toward one of two terminal states: COMPLETE, representing successful code generation and validation, or FAILED, triggered when iteration thresholds are exceeded or unrecoverable logic errors are detected.

Overall, this pipeline state schema functions as the persistent memory of the system, maintaining all key information including the original task description, robot and object specifications, intermediate plans, validation feedback, and iteration counters. This ensures context preservation, auditability, and state recovery across the entire execution cycle. Through this schema, all agents operate in a synchronized yet modular fashion, each independently responsible for its reasoning scope while collectively contributing to coherent system-level decision-making.

**Performance Validation and Evaluation Limitations** To validate the Self Corrective-Agent Architecture's effectiveness, the authors evaluated it using the same 36-task dataset from the original SMART-LLM study (Kannan et al. 2024) in the AI2-THOR simulation environment. This dataset choice enabled direct assessment of whether our architectural improvements addressed the challenges identified during replication while maintaining comparability with the established baseline. Similar to SMART-LLM, Success Rate (SR), Task Completion Rate (TCR), Goal Condition Recall (GCR), Executability (EXE), and Robot Utilization (RU) were used as the evaluation metrics. MRTA performance is evaluated using five complementary metrics. Executability (Exe) measures the fraction of actions in the task plan that can be successfully executed, validating syntactic and semantic correctness regardless of task completion. Goal Condition Recall (GCR) quantifies task completion accuracy as the ratio of satisfied goal conditions to total required conditions by comparing final achieved states against ground truth. Task Completion Rate (TCR) is binary: it equals 1.0 when GCR = 1.0 (all goals satisfied) and 0 otherwise. Robot Utilization (RU) evaluates multi-robot coordination efficiency by comparing experimental robot transitions against ground truth values, where transitions occur when one robot group finishes and another begins. RU = 1.0 indicates optimal parallelization (matching ground truth), RU = 0 indicates fully sequential execution (transitions equal sub-task count), and intermediate values reflect partial parallelization. Success Rate (SR) is the most stringent metric, equaling 1.0 only when both GCR = 1.0 and RU = 1.0, indicating both perfect task completion and optimal robot utilization. To better quantify and describe the 36 benchmark tasks, a simple categorization was used. The 36 tasks were classified into Elemental, Simple, Compound, and Complex tasks according to tasks are classified based on the complexity of coordination required and the degree of robot heterogeneity. The classification progresses from tasks requiring minimal coordination with homogeneous capabilities to those demanding strategic team formation where individual robots lack sufficient skills or properties to complete sub-tasks independently, necessitating collaborative execution to leverage combined capabilities. Elemental tasks are single-action tasks performed by one robot with all necessary skills, e.g., "Make the kitchen dark". Simple tasks involve multiple objects with homogeneous robots executing sub-tasks either sequentially or in parallel, e.g., "Put apple in fridge and switch off the light". Compound tasks use heterogeneous robots with specialized skills where each robot independently completes assigned sub-tasks, e.g., "Cook the potato and put it in the Fridge". Complex tasks require team formation where robots must collaborate on the same sub-task due to skill or property constraints, e.g., "Toast a slice of the breadloaf".

**TABLE 1.** Baseline Performance Evaluation of SMART-LLM Framework

|  | SR | TCR | GCR | EXE | RU | Tasks Not Executed |
|---|---|---|---|---|---|---|
| Elemental | 0.56 | 0.56 | 0.66 | 0.66 | 0.66 | ** |
| Simple | 0.31 | 0.31 | 0.31 | 0.71 | 0.62 | ***** |
| Compound | 0.10 | 0.17 | 0.32 | 0.58 | 0.57 | ****** |
| Complex | 0.18 | 0.18 | 0.18 | 0.39 | 0.56 | ****** |

*Note.* Each asterisk (*) represents one task that was not run in that category.

Table 1 presents the baseline performance metrics from the SMART-LLM replication study across four task complexity categories. The framework achieved success rates ranging from 0.10 (Compound) to 0.56 (Elemental), with executability ranging from 0.39 to 0.71. The asterisks indicate categories where certain tasks failed to execute entirely. These baseline results established the performance benchmark and confirmed the need for architectural improvements.

**TABLE 2.** Evaluation of Self Corrective-Agent Implemention for different categories of tasks.

|  | SR | TCR | GCR | EXE | RU |
|---|---|---|---|---|---|
| Elemental | 0.5 | 0.5 | 0.58 | 0.75 | 1 |
| Simple | 0.25 | 0.25 | 0.63 | 0.83 | 0.75 |
| Compound | 0.21 | 0.28 | 0.55 | 0.88 | 0.82 |
| Complex | 0.5 | 0.63 | 0.63 | 0.86 | 0.63 |

As illustrated from table 2 The framework achieved success rates ranging from 21% (compound tasks) to 50% (elemental and complex tasks). It generally has a better performance for Elemental tasks, due to fact that Compound tasks contain the highest number of test cases and involve longer multi-step action chains, making them more susceptible to partial goal completion even when execution succeeds. However, it was noted that the code generated by the SMART-LLM agent could be incompatible with the simulation environment and the robot agent's motion planning. With such incompatibility, the system reports the code as "not executable" by having a low Exec metric and reduces the whole task allocation success rate. After thorough inspections, it was identified that such errors happen due to the lacked training and fine-tuning of the code generation part. In addition, the AI2-THOR's object state detection mechanisms and spatial tolerance thresholds were inconsistent with the code generation LLM, which could also be the reasons. When trying to fix the problem, nonetheless, the SMART-LLM agent will repeat the whole task allocation process from task decomposition, which uses a very large model and wastes computational resources.

Overall, despite these evaluation limitations, the replication study confirmed three critical findings: (1) LLM-driven coordination is feasible for multi-robot task allocation, (2) structured validation mechanisms significantly improve reasoning reliability, and (3) hierarchical retry systems effectively handle LLM output inconsistencies.

With such LLM-MRTA feasibility established, the authors developed the LTAA framework to improve the performance of LLM-MRTA. LTAA evaluation uses the TEACh dataset the standard benchmark employed for Q-learning, DQN, and DP comparisons in MRTA research. This dataset shift from AI2-THOR household tasks to construction robotics scenarios enables direct performance comparison with established algorithmic baselines under identical evaluation conditions.

The LTAA framework specializes the allocation component from the Self Corrective-Agent Architecture, accepting pre-decomposed tasks (as traditional algorithms do) and focusing exclusively on robot-task assignment decisions. This scope alignment ensures that LLM-based and algorithm-based approaches are evaluated on equivalent problem formulations.

**LangGraph Task Allocation Agent Framework Architecture**

Building on the self-corrective principles established above, the LTAA framework implements a nine-node workflow. Unlike the Self Corrective-Agent Architecture which handled full-pipeline coordination, LTAA framework focuses exclusively on the allocation decision process, accepting pre-decomposed tasks and producing robot-task assignments comparable to traditional algorithmic outputs.

The LTAA framework introduces three core technical innovations that enable competitive performance with traditional optimization algorithms while maintaining reasoning transparency: (1) *phase-adaptive allocation strategies* that dynamically adjust decision priorities throughout project progression, (2) *multi-stage validation with hierarchical retry and self-correction mechanisms* ensuring LLM reasoning quality, and (3) *context-aware reasoning integration* that structures allocation decisions with quantitative capability priors. These contributions are implemented through a nine-node LangGraph workflow that processes each task through structured reasoning stages. The following subsections detail each technical contribution, followed by complete workflow integration.

Figure 5 presents the complete nine-node workflow architecture, with detailed technical contributions described in subsequent subsections.

**Phase-Adaptive Allocation Strategy**

Traditional multi-robot task allocation methods apply static optimization objectives throughout task sequences, treating all allocation decisions with uniform priorities. This approach assumes that optimal strategies evolve throughout project progression. Early-phase allocation requires an emphasis on execution feasibility to establish operational baseline, while late-phase allocation must prioritize workload equity as opportunities for rebalancing diminish. The authors address this limitation through phase-adaptive allocation strategies that systematically adjust the trade-off between success rate maximization and workload balance based on project progression.

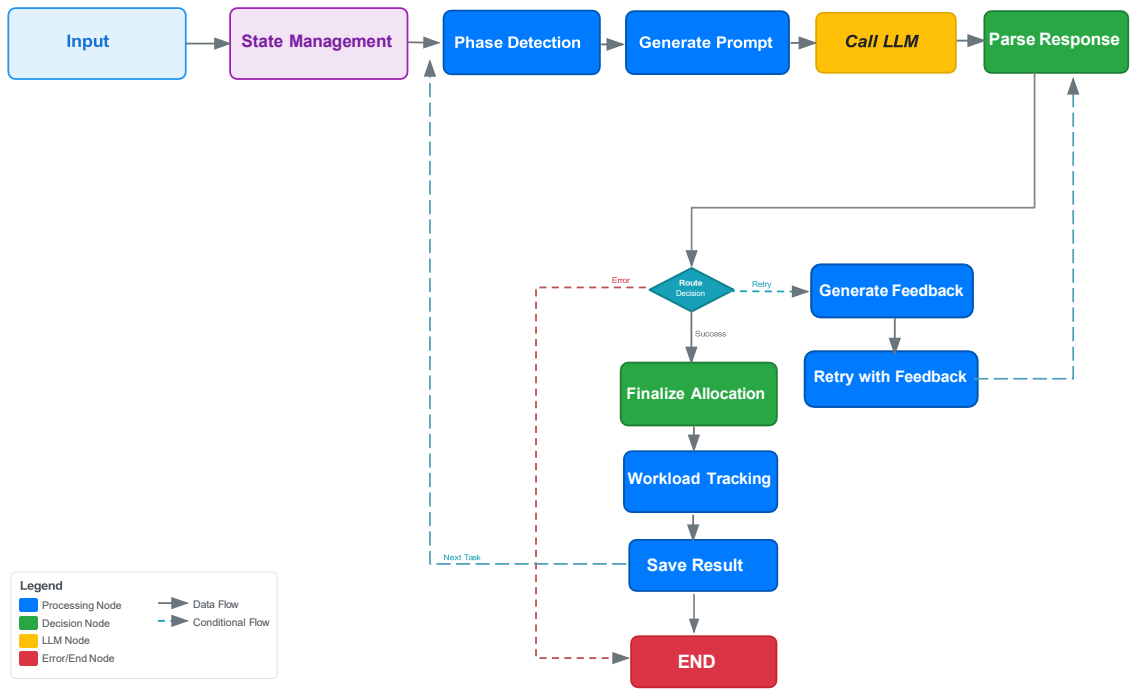

**Fig. 5.** LangGraph based Task Allocation Agent Framework

**Temporal Phase Classification Framework** The introduction of phase detection in this framework draws on prior research highlighting the temporal adaptive nature of multi-robot allocation. Choudhury et al. (2024) introduced MRTA as a dynamic decision-making process that evolves with environmental and operational context, requiring continual adjustment of allocation strategies. Nunes et al. (2017) demonstrated that task characteristics and progression inherently divide the allocation process into functional stages, where each stage demands context-specific reasoning. Building on these insights, the proposed framework formalizes this temporal evolution through an explicit phase-detection mechanism that enables reasoning strategies to adapt systematically as the task sequence advances.

In this framework, phase detection node divides the allocation process into three temporal stages: early, middle, and late to represent the evolving decision priorities and the project stage assigned. The phases are defined proportionally to task completion, motivated by prior research highlighting temporal evolution in MRTA (Choudhury et al. 2024, Nunes et al. 2017). The early stage (0–33%) emphasizes maximizing success rates when flexibility is highest; the middle stage (34–66%) focuses on maintaining workload balance as disparities begin to emerge; and the late stage (67–100%) concentrates on stabilization and equitable completion. This tripartite segmentation, derived from the 36-task sequential structure, provides a clear temporal logic that balances interpretability with adaptive control throughout the allocation process.

**Phase Detection Algorithm** The framework implements automated phase detection through continuous monitoring of allocation progress, formalizing temporal boundaries and strategic weight assignment:

---

**Algorithm 1. Phase Detection for Allocation Workflow**

1. For each allocation cycle $i$ in task sequence $T$:

    (a) Compute total allocated tasks: $c_i$
    (b) Compute total tasks: $C$
    (c) If $C = 0$, set phase ← "early"
    (d) Else compute progress ratio: $p = c_i/C$
    (e) If $p < 0.33$, set phase ← "early"
    (f) Else if $0.33 \leq p < 0.67$, set phase ← "middle"
    (g) Else, set phase ← "late"
    (h) Update workload indicators:
        i. balance_urgency ← $p$       (increases with progress)
        ii. allocation_phase ← phase
    (i) Return phase and balance_urgency to the state tracker.

---

**Fig. 6.** Phase Detection Algorithm used in the LTAA Framework.

The algorithm executes at each allocation cycle (i.e., before each task assignment), computing the progress ratio and determining the current operational phase. Strategic weights ($\alpha_{success}$, $\alpha_{balance}$) define the relative importance of success rate maximization versus workload balance in LLM reasoning guidance To operationalize these strategic weights, the framework requires quantitative assessment of the current workload distribution specifically, how far the system has deviated from ideal balance. The following workload balance quantification framework provides this essential measurement capability.

**Workload Balance Quantification** The workload balancing logic in this framework is conceptually aligned with the competency adjustment and workload balancing principles introduced by Lee et al. (2018), where robot load distribution is continuously evaluated to prevent over- or under-utilization. In their approach, workload is defined as the ratio of assigned tasks to the total available capacity of each robot, and balancing is achieved by reallocating tasks whenever deviations exceed acceptable thresholds. This provides a strong theoretical basis for the balance-tracking variables used in the present system.

Each robot's workload deviation from the ideal target $W_{target} = \frac{36}{3} = 12$ is expressed as:

$$\delta_r = W_r - W_{target}. \tag{1}$$

Where,

$\delta_r$ Deviation of robot $r$'s workload from the target.

$W_r$ is the number of tasks assigned to robot $r$.

The balance score $B$ is then calculated as: Balance score equation:

$$B = 100 \times \left(1 - \frac{\max_r |\delta_r|}{W_{target}}\right) \tag{2}$$

Where,

$B$ Balance score, representing workload fairness (0–100 scale).

$\max_r |\delta_r|$ The maximum absolute deviation among all robots, i.e., the robot that is currently

farthest from the ideal workload.

It measures the worst-case imbalance in the system.

Thus, the higher the deviation $\max_r |\delta_r|$, the smaller the balance score $B$.

A perfectly balanced distribution yields $\delta_r = 0$ for all robots, leading to $B = 100\%$, which normalizes fairness on a 0–100 scale.

A perfectly balanced workload yields $\delta_r = 0$, $B = 100$, while larger deviations reduce the score.

Building upon this foundation, the framework classifies imbalance severity into four discrete levels: low, moderate, high, and critical based on the ratio between the observed workload deviation and the target load. This multi-tier severity structure aligns with MRTA literature, where deviation or urgency thresholds are used to distinguish mild, moderate, severe, and critical conditions that trigger different reallocation behaviors (Choudhury et al. 2022, Lee et al. 2018, Faruq et al. 2018). This classification enables the framework to implement a balanced strategy to each allocation phase. For instance, low or moderate imbalance triggers only monitoring and gradual correction, whereas high or critical imbalance invokes immediate adjustment during the next task allocation cycle. By coupling these severity levels with phase-specific priority, early, the system maintains a balanced progression of task assignments while avoiding abrupt redistributions that could destabilize the allocation sequence.

The numerical thresholds defining each imbalance level are logically derived from the problem scale and robot distribution. Given a total of 36 tasks allocated among three robots, the ideal target workload per robot is 12 tasks. A deviation of up to $\pm 1$ task (25% of the target) is therefore considered a minor imbalance, representing operational tolerance without performance degradation. Larger deviations promote increased inequality in task distribution and are thus classified as moderate, high, or critical imbalances. This proportional approach ensures that severity levels remain interpretable and scalable across different task volumes.

**Example Calculating Balance Score**

Suppose at a mid-stage of allocation the task distribution is:

Here, the maximum deviation $\delta_{max} = 7$.

**TABLE 3.** Example of Calculating Balance Score

| Light | 19 | +7 |
|---|---|---|
| Medium | 10 | -2 |
| Heavy | 7 | -5 |

Substituting into the formula:

$$B = 100 \times \left(1 - \frac{7}{12}\right) = 100 \times (1 - 0.5833) = 41.6.$$

Thus, the Balance Score = 41.6%, representing high imbalance.

According to the severity classification:

- Low imbalance: ≤ 3 tasks deviation

- Moderate: 3–6 tasks

- High: 6–12 tasks

- Critical: > 12 tasks

This scenario would trigger balance-prioritized behavior in later allocation phases, encouraging the LLM to redirect tasks toward underutilized robots even if Heavy's success probability slightly decreases.

**Phase-Specific Strategic Framework** The Table 4 formalizes the decision philosophy governing each allocation phase:

The temporal phase classification, workload balance quantification, and phase-specific strategic frameworks described above are integrated into the workflow through Node 1 (Phase Detection), which serves as the initial reasoning checkpoint in the nine-node LTAA architecture.

This node serves as the initial reasoning checkpoint, receiving system state and computing temporal context for downstream reasoning stages. The Phase Detection Node serves as the initial reasoning checkpoint in the LTAA framework, where the framework evaluates task progress and updates workload-related metrics. As shown in Figure 9a, this node receives four key state variables from the system: the number of completed and total tasks, the current workload distribution across robots, and the ideal target workload. These parameters collectively define the allocation context at each iteration of the process.

TABLE 4. Phase-Adaptive Allocation

| Phase | Philosophy | Strategic Weight (Success: Balance) | Decision Priorities and Behavioral Adaptation |
|---|---|---|---|
| **Early Phase (0–33%) – Foundation Building** | "Success first, balance later." Focus on achieving initial task success to stabilize the system. | 80%: 20% | Prioritizes successful task completions to build a strong foundation. Accepts moderate workload imbalances to minimize early-stage failures. |
| **Middle Phase (34–66%) – Strategic Balancing** | "Sustainable performance with growing fairness awareness." Introduces fairness considerations without compromising stability. | 60%: 40% | Shifts focus to balancing task success with workload distribution. Employs more nuanced decision logic to evaluate acceptable performance–fairness trade-offs. |
| **Late Phase (67–100%) – Equity Prioritization** | "Finishing fairly is as important as finishing successfully." Emphasizes fairness and equity toward the project's end. | 40%: 60% | Recognizes diminishing opportunities to rebalance workloads. Favors equitable distribution of remaining tasks even at the cost of slight success-rate reductions. |

Within this node, the framework computes the current progress ratio and uses it to determine the system's operational phase classification as early, middle, or late. Each phase represents a distinct decision context, guiding how subsequent reasoning nodes balance success rates and workload fairness. The node then calculates the balance score to quantify how evenly tasks are currently distributed and identify imbalance severity based on deviation from the ideal workload. These metrics encapsulate both temporal progress and workload dynamics in a compact form.

The node concludes by updating the system state with the derived parameter: allocation_phase, balance_score, imbalance_severity, and operational_mode which are passed forward to the Prompt

Generation Node. This ensures that all downstream reasoning stages operate with up-to-date contextual awareness of progress and workload status. As illustrated in Figure 9a, this process forms a unidirectional flow where the Phase Detection Node transforms raw operational data into structured decision context for the subsequent reasoning phase.

**Multi-Stage Validation with Hierarchical Retry Mechanisms**

The phase-adaptive allocation strategy requires reliable LLM reasoning to make context-appropriate decisions. However, LLM outputs exhibit inherent variability and may fail to meet quality standards for safety-critical construction applications. To address the generation inconsistencies identified during SMART-LLM implementation Section , the authors developed a comprehensive validation framework that ensures reasoning quality through weighted multi-criteria assessment and structured feedback mechanisms.

The parsing system extracts structured allocation decisions from LLM markdown responses using regex pattern matching. This pattern-based text parsing technique identifies specific formatted sections within the LLM's markdown output (e.g., allocation decisions, success percentages, reasoning explanations) and converts them into structured data for validation. A comprehensive validation system applies eight validation rules including elimination of invalid allocations, trade-off analysis verification, workload awareness checking, and logical consistency evaluation.

The adoption of the weighted validation framework was driven by the limitations observed in conventional rule-based validation systems, which often fail to capture the enhanced reasoning quality and contextual awareness essential for LLM-driven task allocation. Recent studies on LLM evaluation (Guo et al. 2023, Evaluation and Benchmarking of LLM Agents, 2025) emphasize that multi-aspect, criterion-based assessment provides a more reliable measure of model reasoning performance than binary correctness checks. Building on these insights, the proposed validation system employs a weighted scoring approach that evaluates reasoning outputs across multiple dimensions like explanation clarity, success-rate consistency, workload awareness, trade-off justification, phase compliance, and confidence alignment reflecting both the technical soundness and the interpretive quality of the model's decisions. Each dimension contributes proportionally to a composite quality score, with higher weights assigned to factors that more directly influence allocation outcomes, such as reasoning quality and success-rate accuracy, while supporting aspects like confidence justification receive lower weights. This configuration ensures that the validation process remains both performance-sensitive and context-aware, effectively balancing interpretability and operational rigor in evaluating the model's reasoning behavior.

**Weighted Multi-Criteria Validation:** The validation framework employs eight criteria with performance-based weights. Following G-Eval's weighted summation approach for LLM evaluation by Liu et al. (2023), the framework applies criterion-specific weights that aggregate into an overall quality score. Weight Assignment Criteria were organized into four importance tiers based on their contribution to allocation correctness. The weight structure follows established multi-criteria evaluation principles: criteria judged to be of equal importance are assigned equal weights, consistent with standard additive value modeling approaches described by Pöyhönen and Hämäläinen (2001). The tier distribution allocates 40% to critical dimensions, 30% to operational coordination, 20% to contextual factors, and 10% to verification checks, with each tier containing two equally weighted criteria:

**Critical Dimensions (0.20 each, 40% combined)**

1. Explanation Quality (Weight: 0.20): Reasoning depth and clarity.

2. Success Rate Accuracy (Weight: 0.20): Alignment between predicted and actual success probabilities.

**Operational Coordination (0.15 each, 30% combined)**

3. Trade-off Analysis (Weight: 0.15): Explicit consideration of competing objectives.

4. Workload Awareness (Weight: 0.15): Consideration of current task distribution.

**Contextual Factors (0.10 each, 20% combined)**

5. Mode Compliance (Weight: 0.10): Adherence to operational guidelines.

6. Phase Consistency (Weight: 0.10): Appropriate phase-specific strategies.

**Verification Checks (0.05 each, 10% combined)**

7. Logical Consistency (Weight: 0.05): Internal reasoning coherence.

8. Confidence Justification (Weight: 0.05): Appropriate articulation of confidence levels.

The specific weights (0.20, 0.15, 0.10, 0.05) follow a 4:3:2:1 importance ratio across tiers. This graduated structure ensures no single criterion dominates the evaluation (maximum weight 20%) while maintaining clear priority differentiation and balanced representation of each dimension.

$$Q = \sum_{i=1}^{8} w_i q_i \quad \text{and} \quad \sum_{i=1}^{8} w_i = 1 \tag{3}$$

where $q_i \in [0, 1]$ is the rule-level score and $w_i$ its importance.

Acceptance: $Q \geq 0.6$ for validation approval.

The validation acceptance threshold of $Q \geq 0.6$ follows established methodologies for LLM evaluation metrics, where thresholds are determined based on confidence levels and risk tolerance to ensure outputs meet necessary standards for reliability Sarmah et al. (2024).

The validation system calculates overall quality scores using weighted averages of individual rule assessments. Quality scores range from 0.0 (perfect failure) to 1.0 (perfect reasoning), with acceptance thresholds typically set between 0.6-0.8 depending on application requirements.

The weighted validation framework operates in two modes: when reasoning quality meets the acceptance threshold ($Q \geq 0.6$), the allocation proceeds to finalization; when quality falls below threshold ($Q < 0.6$), the framework invokes hierarchical retry mechanisms to iteratively improve reasoning quality through structured feedback.

**Hierarchical Retry Escalation Strategy** When validation identifies reasoning deficiencies (Q<0.6), the framework implements a three-tier retry escalation mechanism with progressively intensive guidance:

**Tier 1: Validation-Based Retry with Specific Feedback**

When LLM reasoning fails quality standards, the system analyzes specific deficiencies across eight quality criteria and generates targeted feedback. The first retry combines original prompts with specific improvement guidance, typically resolving 60–70% of validation failures.

**Tier 2: Progressive Enhancement with Comprehensive Guidance**

If the first retry fails, the system provides comprehensive improvement guidance with concrete examples of better reasoning. Enhanced prompts include detailed reasoning templates and examples, addressing approximately 80–90% of remaining validation failures.

**Tier 3: Final Attempt with Maximum Support**

The final retry combines all previous feedback with step-by-step reasoning frameworks and structured templates. This intensive mentoring typically yields detailed outlines for constructing appropriate responses.

**Fallback Allocation Strategy**

When all three retry tiers fail, the framework implements a conservative fallback allocation strategy based on established principles of graceful degradation in robotic systems (Silva et al. 2024). Research demonstrates that fail-safe systems require "conservative bounds" and default behaviors to maintain operational continuity when primary mechanisms fail Porges et al. (2021). The conservative allocation (typically Light Robot with 50% success estimate) follows established practices for task acceptance under resource constraints, ensuring system functionality despite allocation failures Rehman et al. (2022).

**Retry Escalation:**

Attempt 1: Original prompt + specific validation feedback

Attempt 2: Enhanced prompt + comprehensive guidance + examples

Attempt 3: Maximum support + reasoning templates + structured framework

Fallback: Conservative allocation + detailed failure logging

**Validation and Feedback Loop (Nodes 4-6)** The validation and retry mechanisms are implemented through three interconnected nodes (Nodes 4, 5, 6) that form a feedback loop ensuring reasoning quality:

**Node 4: Response Parsing and Validation** As shown in Figure 9b the system receives as input the raw natural-language reasoning output generated by the LLM in the previous stage. This response is parsed into structured components like allocation decision, expected success rate, reasoning explanation, confidence level, and trade-off justification which form the basis for validation. The parsed content is then evaluated through a weighted multi-criterion validation system that assesses reasoning quality, success-rate consistency, workload-driven awareness, trade-off analysis, and operational compliance. Each rule contributes to an aggregated quality score, determining whether the response is accepted or routed for feedback-driven correction. The node outputs a fully validated allocation record with its corresponding quality score and validation status, which is then passed to the subsequent stage for finalization and workload tracking.

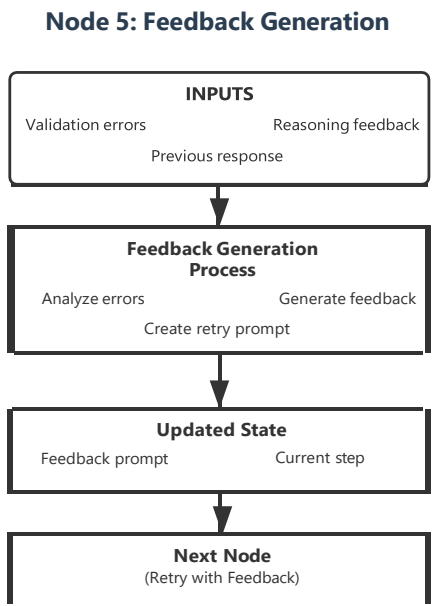

**Fig. 7.** Flow chart of Node 5

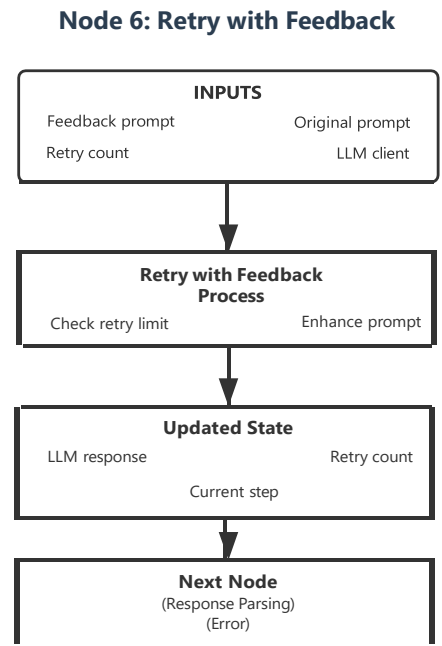

**Fig. 8.** Flow chart of Node 6

**Node 5: Response Parsing** In this node, Figure 9b the framework receives as input the validation errors and reasoning feedback detected in the preceding stage when a response fails one or more validation criteria. Using this information, the node constructs a structured feedback prompt that explicitly lists detected errors, corresponding rule violations, and improvement recommendations.

The prompt sets all targeted guidance for the subsequent retry process. The node's output is an enriched feedback message appended to the existing task state, ensuring that all diagnostic information such as error type, affected reasoning criteria, and specific corrective recommendations is preserved for the next node in the workflow. This design enables precise and context-aware correction without altering the task's original reasoning context.

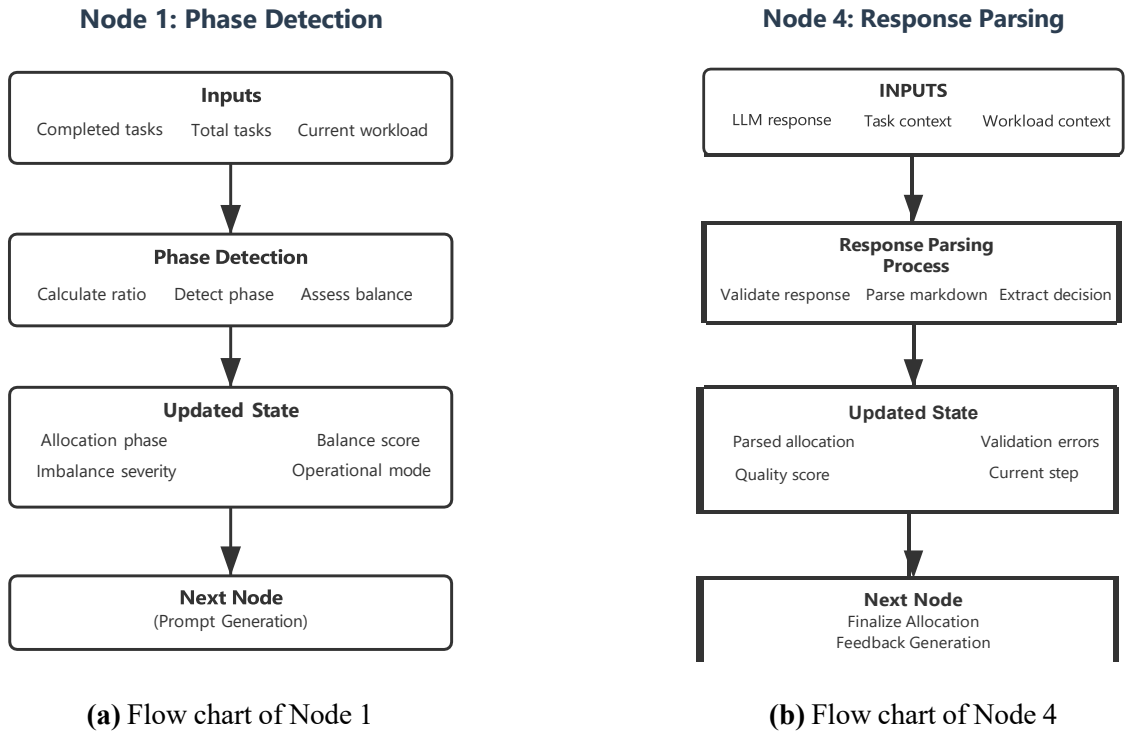

(a) Flow chart of Node 1  (b) Flow chart of Node 4

**Fig. 9.** Flow charts of Node 1 and Node 4

**Node 6: Retry with Feedback Integration** This node manages retry logic with a maximum of three attempts per task. Enhanced prompts combine original context with specific feedback to guide improved reasoning on re-attempts.as illustrated in the Figure 8

**Context-Aware Reasoning Integration**

The phase-adaptive strategies are implemented through sophisticated LLM integration that translates strategic weights into contextual reasoning guidance. The prompt engineering system dynamically incorporates phase-specific priorities and natural language instructions, ensuring the LLM applies appropriate trade-off logic at each project stage. For instance, during early phase allocation, prompts emphasize success rate maximization with phrases like "prioritize robots with highest success rates" and "accept moderate workload imbalances," while late phase prompts shift to "favor workload equity" and "consider balance as primary factor." This adaptive prompt mechanism enables the LLM to naturally implement apply the strategic frameworks defined in the phase-adaptive approach.

The authors operationalize this integration through three mechanisms: (1) probabilistic robot capability modeling that computes task-specific success rates, (2) structured prompt generation that embeds quantitative priors and workload context, and (3) standardized response formatting that enables reliable parsing and validation.

**Probabilistic Robot Capability Modeling**

Three types of robots with Features Light, Medium, and Heavy are modeled with distinct strengths and probabilistic success profiles. The success profiles for each robot are derived from probabilistic associations between task features and robot capabilities. This approach follows the formulation of Faruq et al. (2018), where task completion likelihoods are modeled as expected success probabilities based on the uncertainty of feature capability alignment. Accordingly, each robot's overall success rate for a given task is obtained by aggregating its feature-conditioned success values.

Each robot's ability to perform a given task depends on the overlap between the task's required features and the robot's capability distribution. Tasks are sequentially processed through a queue initialized with all 36 tasks from the TEACh dataset (Padmakumar et al. 2022). Here sequential processing means where each task is individually passed through the full LangGraph workflow including phase detection, prompt generation, reasoning, validation, and allocation before proceeding to the next one. Upon completion of each task, the state updates progress metrics and queue length until all tasks are successfully assigned.

For each task $t$ defined by a set of required features $F_t = \{f_1, f_2, ..., f_n\}$, the framework computes an aggregate success rate $S(r, t)$ for each robot $r$ using weighted averages. Here, the success contribution of each feature is represented as the product of the feature's success likelihood and the robot's corresponding capability weight ($k \times r$), reflecting the joint probability that a robot with certain skills successfully executes a feature-dependent task.

**TABLE 5.** Robot Specifications and Optimal Applications

| Robot | Specialization | Optimal Applications |
|---|---|---|
| Light | Precision Specialist | Assembly, inspection, delicate manipulation |
| Medium | Balanced Generalist | Mixed operations and general-purpose tasks |
| Heavy | Force Specialist | Heavy lifting, material handling, construction |

$$S(r, t) = \frac{1}{|F_t|} \sum_{f \in F_t} M[f][r] \tag{4}$$

where $M[f][r]$ is the success probability of robot $r$ on feature $f$. The following example illustrates how robot capability definitions and success matrices are structured within the framework. It demonstrates how each robot's attributes, such as skill specialization and category, are encoded alongside the corresponding success probabilities for different task features. This representation provides the foundation for calculating the aggregated success scores discussed earlier and enables direct mapping between feature requirements and robot performance characteristics.

**Example 1: Single-Feature Task**

Task: "Stop"

Features: [dexterous]

**TABLE 6.** Example of Single-Feature Task

| Robot | Success Rate for "dexterous" |
|---|---|
| Light | 0.8 |
| Medium | 0.6 |
| Heavy | 0.4 |

Thus, for each robot:

$$S_{Light} = 0.8, \quad S_{Medium} = 0.6, \quad S_{Heavy} = 0.4.$$

Since this task involves a single feature, the aggregate success rate equals the base feature probability.

These computed success rates ($S(r, t)$ for each robot-task pair) serve as quantitative priors that inform the prompt generation process, providing the LLM with concrete performance expectations for allocation reasoning.

**Context-Aware Prompt Generation Framework**

1. **Task specification**. Action ID, name, type, and required features (e.g., heavy, dexterous, careful) are presented with short, contextual definitions to anchor the LLM's reasoning.

2. **Robot Capability Assessment**. This component evaluates each robot's suitability for the specific task by calculating success rates using Table 8. The assessment provides the LLM with quantitative success probabilities for each robot-task pairing, enabling informed comparison of robot performance capabilities. For example, for a dexterous task, the assessment might show: Light Robot (80% success), Medium Robot (60% success), and Heavy Robot (40% success), giving the LLM clear performance differentials to incorporate into its allocation reasoning.

3. **Workload context**. In this stage, the framework dynamically generates the user prompt by embedding the latest workload and fairness data into a predefined reasoning template. The system prompt remains constant, defining the model's general reasoning behavior, while the user prompt is built in real time using current task features, per-robot task counts, target workload (12 per robot), deviations ($\delta_r$), and the computed BalanceScore. This ensures the model remains aware of workload disparities and maintains fairness during allocation.

4. **Allocation Stage Guidance**. The system evaluates the fundamental trade-off between task success maximization and workload fairness when robots with higher success rates also have heavier current workloads. Trade-offs are evaluated through quantitative assessment of success rate gaps, workload deviations, and phase-specific strategic weights that systematically resolve competing priorities. Detailed examples of trade-off evaluations are provided in Appendix 2.

**Node 2: Context-Aware Prompt Generation**

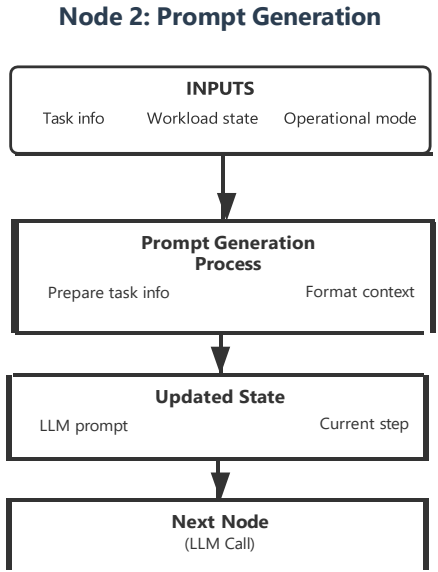
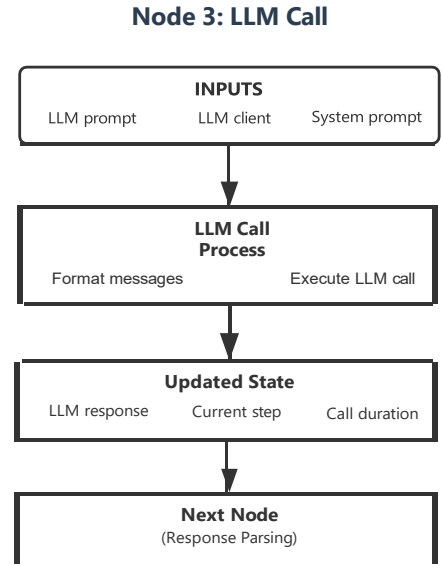

**Fig. 10.** Flow chart of Node 2             **Fig. 11.** Flow chart of Node 3

In this Node, the framework constructs a contextually rich user prompt that integrates all relevant task, workload, and fairness information before invoking LLM reasoning. As illustrated in Figure 10, this node receives as input the current task details, phase state, and workload distribution parameters from the previous node. It dynamically embeds these values such as per-robot task counts, balance score, deviations ($\delta_r$), and success priors into a predefined reasoning template that complements a constant system prompt defining the model's behavioral intent. The resulting prompt explicitly states the current phase and reasoning objective (e.g., prioritizing success or balancing fairness) and organizes the output in a structured markdown format containing per-robot analysis, success priors, a single allocation decision, justification, post-allocation workload, confidence level, and trade-off summary. This ensures that each reasoning cycle is context-aware, temporally aligned, and consistent with the evolving state of task allocation, providing a smooth transition to the subsequent LLM reasoning node.

Before executing LLM reasoning, the authors establish standardized response formatting requirements to enable reliable parsing and validation of allocation decisions.

**Response Format Standardization**

The structured response framework addresses a critical information processing challenge identified in recent LLM research. Industry studies demonstrate that standardized output formats are essential for incorporating LLMs into production workflows, as format inconsistencies complicate parsing and undermine system reliability (Liu et al. 2024, Xia et al. 2024). This standardization is based on empirical evidence from industry professionals who identified specific requirements for structured LLM output to achieve downstream processing accuracy and reduce bias consumption waste (Tam et al. 2024). The framework's design follows established principles for LLM output constraint and balances generation quality with format compliance, enabling reliable integration into multi-robot task allocation systems.

This framework enforces structured markdown response formats that include:

- Robot analysis with success rates and workload status

- Explicit allocation decision with expected success percentage

- Detailed reasoning explanation (2-4 sentences)

- Post-allocation workload projection

- Confidence level assessment (High/Medium/Low)

- Trade-off summary detailing sacrifices made

Context-aware reasoning is implemented through Nodes 2 and 3, which generate structured prompts and execute LLM inference: Node 2 (Context-Aware Prompt Generation) is detailed above. Node 3 executes LLM reasoning:

**Node 3: LLM Reasoning Execution** In this node as shown in Figure 11, the framework executes the reasoning process by invoking the LLM using a structured combination of two prompt components: the system prompt, which defines the agent's overall reasoning behavior and decision policies, and the user prompt, dynamically generated in the previous node to reflect the current allocation context. These prompts are concatenated sequentially to form the model input where the system prompt establishes global guidance, and the user prompt provides task-specific situational data such as success priors, workload deviations, and phase context. The call is executed through the invoke() function of the LLM client, which transmits the combined prompt as a structured message sequence. The LLM then produces a natural language response containing its allocation reasoning, decision, and confidence level, which is returned for parsing and validation in the subsequent node.

Here, the system receives as input the combined prompts prepared in the previous stage comprising the static system prompt, which encapsulates the reasoning policy and operational rules, and the dynamically generated user prompt, which embeds the LLM-specific features, workload state, and phase information. These components are integrated into a structured message sequence and passed to the LLM client for inference. During execution, the model interprets this contextual input to produce natural-language reasoning containing an allocation decision, explanatory justification, expected success rate, and confidence level. This response serves as the node's output and forms the direct input for the reasoning parsing and validation stage, ensuring continuity and traceability throughout the workflow.

**Complete Workflow Integration and State Management**

**Integrated Nine-Node Architecture** The system employs multiple reasoning types including phase-adaptive reasoning that adjusts priorities based on project completion (early phase prioritizes success, late phase emphasizes workload balance), contextual trade-off analysis that weighs success rates against workload fairness, and validation-based iterative improvement that provides feedback for reasoning enhancement. LangGraph, a state management and workflow orchestration library, enables the creation of complex multi-step reasoning processes with conditional routing and error recovery mechanisms.

For example, when assigning a dexterous task, the system first detects the current phase (early/middle/late), generates a comprehensive prompt including robot success rates (Light) Robot:

80%, Medium Robot: 60%, Heavy Robot: 40%) and current workload status, sends this to the LLM for reasoning, validates the response quality, and either finalizes the allocation or provides feedback for retry. The LLM might reason: "Choose Light Robot despite having 2 more tasks than others because the 20% success advantage justifies the minor imbalance in early phase, but this decision will require balancing in later allocations."

The workflow architecture consists of three primary computational layers:

1. **Input Processing Layer** Loads tasks from the TEACh dataset (36 tasks with action IDs, names, types, and features), initializes robot definitions with success matrices, establishes TaskAllocationState (the system's memory bank that tracks task queue, robot workloads, decision history, and balance scores throughout the allocation process), and establishes LLM client connections

2. **LangGraph Reasoning Layer**: Executes the 9-node sequential workflow with LLM integration and validation

3. **Output Generation Layer** Produces allocation results with comprehensive quality metrics and reasoning explanations, such as quality scores, success rates, workload distributions and detailed reasoning like "Light Robot chosen for dexterous task due to 20% success advantage (80% vs 60%) over Medium Robot, justified in early phase despite creating minor workload imbalance"

**State Management System**

The framework employs a comprehensive state management system here the state refers to workflow, and they are defined through a Python TypedDict structure that maintains all relevant information throughout the allocation process. These state variables collectively represent the relevant information, including task data, robot capability definitions, success matrices, LLM prompts and responses, validation outcomes, workload balance metrics, and workflow control parameters. The state variables are organized into four key functional categories that represent the core components of the allocation process: task information management, robot capability data, LLM processing variables, and workload balancing metrics.

*Task Information Management*

Each task contains a unique identifier, (eg: "action_id": 0) descriptive metadata (eg: action_name": "Stop", "action_type": "Motion"), and required capability features (e.g., heavy, dexterous, careful).

**Example task structure:**

{"action_id": 0, "action_name": "Stop", "action_type": "Motion", "features": ["dexterous"]}

Tasks are sequentially processed through a queue initialized with all 36 tasks from the TEACH dataset. Here sequential processing means where Each task is individually passed through the full LangGraph workflow including phase detection, prompt generation, reasoning, validation, and allocation before proceeding to the next one. Upon completion of each task, the state updates progress metrics and queue length until all tasks are successfully assigned.

*LLM Processing Variables*

The system records:

- Generated prompts containing contextualized descriptions of the task, robot abilities, and fairness status.

- Raw LLM responses and parsed reasoning outputs.

- Validation scores and feedback from the multi-rule quality control system.

In workflow, these variables establish the reasoning pipeline that transforms input task data into validated allocation decisions. The generated prompts encode the current context, workload state, and success profiles, the LLM responses produce candidate allocations, and the parsed outputs are subsequently verified and integrated into the state for execution. This continuous exchange ensures that language-based reasoning directly informs the quantitative allocation framework in real time.

*Workload Balancing Variables*

Workload balance formulation and imbalance severity classification are detailed in Section Workload Balance Quantification. The state management system continuously tracks balance scores (Equation 2), workload deviations ($\delta_r$), and imbalance severity levels throughout the allocation process.

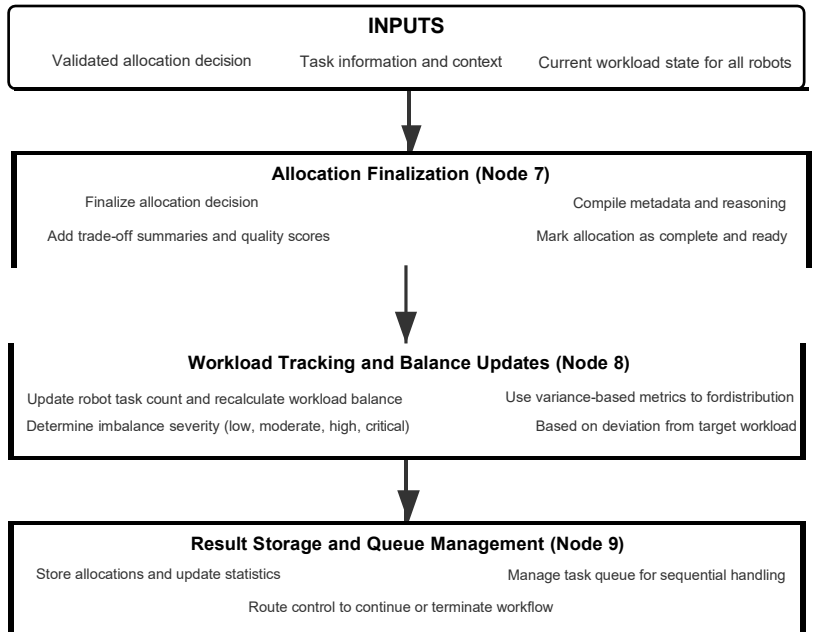

**Fig. 12.** Flow chart of Node 7, Node 8 and Node 9

**Final Workflow Stages: Allocation Finalization and Result Management** The final stages of the LTAA framework Nodes 7, 8, and 9 implement standard operational procedures for allocation finalization, workload tracking, and result management. Unlike the reasoning-intensive processes in Nodes 1-6, these nodes execute deterministic data management operations without complex decision logic. Given their straightforward computational nature and the detailed treatment of core reasoning mechanisms in previous nodes, the authors summarize their functionalities in Table 7 for efficient reference.

Node 7 finalizes validated allocations by consolidating decision metadata, Node 8 updates workload metrics using the balance score framework (Equation 3), and Node 9 manages result storage while routing workflow control. These operations complete the allocation cycle, preparing the system for the next task or workflow termination. Figure 12 illustrates the integrated operation of these three nodes as a unified finalization pipeline, demonstrating their sequential execution and data dependencies within the complete LTAA framework architecture.

**TABLE 7.** Final Stage Workflow Components: Allocation Finalization and State Management

| Description | Node 7: Allocation Finalization | Node 8: Workload Tracking and Balance Updates | Node 9: Result Storage and Queue Management |
|---|---|---|---|
| **Function / Purpose** | Creates the finalized allocation decision after validation, compiling all metadata such as reasoning explanations, confidence levels, trade-off summaries, and quality scores. Marks the allocation as complete and ready for tracking. | Updates robot task count and recalculates workload balance using variance-based metrics. Determines imbalance severity (low, moderate, high, critical) based on deviation from target workload. | Stores completed allocations, updates processing statistics, and manages the task queue for sequential task handling. Routes control to either continue with remaining tasks or terminate the workflow. |
| **Inputs** | 1. Validated allocation decision<br>2. Original task information<br>3. Quality scores and metadata | 1. Final allocation decision<br>2. Current workload counts for all robots<br>3. Target workload per robot | 1. Finalized allocation decision<br>2. List of all completed allocations<br>3. Queue of remaining tasks |
| **Outputs** | 1. Finalized allocation record with metadata<br>2. Ready for workload tracking and saving | 1. Updated workload counts for all robots<br>2. New balance score and severity level<br>3. Data ready for storage | 1. If continuing: Next task routed to Node 1<br>2. If finished: Complete set of all allocations with final status |

## EXPERIMENTAL SETUP

**SMART-LLM Feasibility Assessment**

To assess whether LLM-driven reasoning can feasibly address MRTA problems, we first replicated the SMART-LLM framework (Kannan et al. 2024) as a baseline feasibility study. This replication served to identify fundamental challenges in LLM-based coordination before developing our enhanced framework.

**Dataset and Environment** The evaluation employed the SMART-LLM benchmark dataset consisting of 36 tasks across four complexity categories: Elemental (single-action tasks), Simple (multiple objects with homogeneous robots), Compound (heterogeneous robots with specialized skills), and Complex (requiring team formation and collaborative execution). Tasks were executed in the AI2-THOR simulation environment.

**LLM Configuration** Claude 4 Sonnet served as the reasoning engine with temperature set to 0.0 to ensure deterministic outputs and reproducible allocation decisions.

**Evaluation Metrics** Following the SMART-LLM methodology, we assessed performance using five metrics: Success Rate (SR), Task Completion Rate (TCR), Goal Condition Recall (GCR), Executability (EXE), and Robot Utilization (RU). These metrics collectively measure both task completion accuracy and multi-robot coordination efficiency.

**Self Corrective-Agent Architecture Validation**

Following the identification of validation inconsistencies and LLM output reliability issues during SMART-LLM replication, we developed and evaluated the Self Corrective-Agent Architecture to demonstrate that structured validation and retry mechanisms could address these challenges and establish LLM feasibility for MRTA applications.

**Framework Parameters** The Self Corrective-Agent Architecture introduced multi-stage validation with hierarchical retry mechanisms. Key parameters included: maximum local retries (up to 3 per agent), validation checkpoints at each stage (decomposition, allocation, code generation), and structured feedback loops for iterative improvement.

The Self Corrective-Agent Architecture was evaluated under identical conditions as SMART-LLM (same dataset, LLM configuration, and metrics) to enable direct architectural comparison.

**LTAA Framework**

*Computational Efficiency Evaluation*

To validate the effectiveness of architectural refinements throughout framework development, the authors evaluate computational efficiency through token consumption and execution time metrics across three framework iterations: SMART-LLM baseline, Cyclic-Agent Architecture, and the proposed LTAA framework.

**Evaluation Protocol** The frameworks differ along two fundamental dimensions impacting computational efficiency. First, coordination scope: SMART-LLM and Self Corrective-Agent implement complete pipelines (task decomposition → allocation → code generation), while LTAA focuses exclusively on allocation of pre-decomposed tasks, aligning with traditional algorithms (Q-learning, DQN) that similarly operate on pre-defined task sets. Second, prompting strategy: SMART-LLM and Self Corrective-Agent employ few-shot learning with concatenated examples from all pipeline stages, introducing substantial input token overhead and context window limitations (Section Framework Design Rationale); LTAA uses dynamic prompt generation, constructing context-aware prompts on-demand without example concatenation (Section Context-Aware Prompt Generation Framework).

SMART-LLM and Self Corrective-Agent were evaluated on 36 household robotics tasks from the AI2-THOR benchmark dataset, maintaining consistency with the original SMART-LLM study. LTAA was evaluated on 36 tasks from the TEACh dataset, reflecting its specialization for construction-specific allocation scenarios and enabling direct comparison with traditional optimization baselines. All frameworks employ Claude-4-Sonnet with temperature 0.1 under identical API conditions to isolate architectural impact on efficiency metrics.

*Metrics*

Computational efficiency is assessed through two complementary metrics:

**Token Consumption**: Total tokens processed (input + output) across all task allocations,

measured via API token counting. This metric directly reflects computational cost, as commercial LLM APIs typically charge per token processed.

**Execution Time**: Total wall-clock time and average time per task required to complete all allocation decisions, measured in seconds. This metric captures end-to-end latency including API call overhead, LLM inference time, and validation processing.

*Performance Benchmarking Against Traditional Algorithms*

With LLM feasibility established through the Self Corrective-Agent Architecture, the LTAA framework evaluation shifted focus to systematic performance benchmarking against traditional optimization algorithms.

**Task Feature Categorization and Robot Capabilities** TEACh actions are embodied, human-like manipulation steps that reflect realistic physical demands such as force exertion, precision handling, and careful manipulation. These natural semantics make the actions directly compatible with capability-based robot models, providing an ideal testbed for feature-driven allocation strategies. Actions in the TEACh dataset are annotated with features reflecting their operational demands, categorized as follows:

- **Heavy tasks**: Require substantial force or endurance (e.g., Pickup, Place, Navigation). Examples include lifting heavy materials or positioning large components.

- **Dexterous tasks**: Demand high precision and fine motor skills (e.g., Break, Slice, Clean). Examples include intricate detailing, sealing joints, or precision assembly.

- **Careful tasks**: Emphasize cautious handling to avoid damage (e.g., Pour, Slice, Fill). Examples include handling fragile materials or aligning precision equipment.

The experimental setup employs three commonly used robot capability types (Chung et al. 2008, Zimmermann et al. 2021, Faruq et al. 2018) with distinct performance profiles, as shown in Table 8. The numbers provided were chosen based on Lightweight robots demonstrate superior performance on careful and dexterous tasks due to their enhanced sensitivity and sub-millimeter repeatability, making them optimal for precision assembly applications (Zimmermann et al. 2021, Agile Robots 2024). Heavy-duty robots excel at force-intensive operations, successfully handling payloads exceeding 300 kg for heavy lifting and material transport tasks (Chung et al. 2008). Medium-payload robots serve as versatile generalists capable of handling diverse task types with moderate performance across different operational requirements (Standard Bots, 2024).

**TABLE 8.** Robots Success Rates

| Robot  | Careful | Dexterous | Heavy |
|--------|---------|-----------|-------|
| Light  | 90%     | 80%       | 50%   |
| Medium | 70%     | 60%       | 70%   |
| Heavy  | 50%     | 40%       | 90%   |

**Allocation Strategy Framework**

During preliminary experiments across the intended runtime without strategic guidance, the LLM consistently exhibited extreme allocation patterns, utilizing only Light and Heavy robots while completely neglecting Medium robots during selection. This behavior-making tendency resulted in suboptimal resource utilization and violated workload distribution principles. The authors choose to test the proposed framework using three operational modes.

The three modes provide explicit strategic frameworks that guide the LLM toward different allocation philosophies: Success-Focused Mode formalizes performance-first decisions, Balanced Mode implements dynamic trade-off reasoning, and Aggressive Balance Mode ensures equity-focused allocation. This structured approach prevents the LLM from spontaneous extreme solutions and enables controlled experimentation with different allocation priorities.

**Success-Focused Mode: Performance-First Philosophy**

Philosophy: Prioritizes mission success above all other considerations, operating under the principle that task completion rates directly determine project viability.

Priority weighting: Task success (90%) vs. Workload balance (10%): The 90:10 split creates a strong performance bias that promotes optimal fairness awareness. The extreme ratio ensures that when robots have significant capability differences, the system consistently selects the most capable option rather than compromising performance for equity.

**Trade-off approach**: Systematically assigns tasks to robots with highest success probabilities, using quantitative success rate comparisons as the primary decision criterion. Workload considerations only influence decisions when success rates are nearly identical.

**Acceptance**: Workload imbalances (prefers robots may receive 2-3× more tasks), potential robot overload from specialized task clustering, and underutilization of less-specialized robots that cannot compete on performance metrics.

**Benefits**: Maximizes overall project success likelihood, optimally utilizes specialized robot capabilities, and minimizes task failure risks through consistent capability-task matching.

**Use case**: Crisis management scenarios where project failure consequences outweigh fairness concerns, time-critical operations, or environments where task success directly impacts safety or mission-critical outcomes.

**2) Balanced Mode: Dynamic Optimization Philosophy**

**Philosophy**: Adapts decision priorities throughout project lifecycle, recognizing that optimal allocation strategies evolve as project context changes and opportunities for correction diminish.

**Priority weighting**: Phase-adaptive (evolves throughout project): Weights shift from 80:20 (success/balance) in early phase to 40:60 in late phase, providing systematic priority evolution that balances immediate performance needs with long-term fairness requirements.

**Trade-off approach**: Implements contextual decision-making that continuously recalibrates based on project progress, current workload distribution, and remaining allocation opportunities. Applies sophisticated reasoning to evaluate competing objectives dynamically.

**Accepts**: Moderate performance reductions in later phases to achieve workload equity, occasional suboptimal task assignments to prevent extreme imbalances, and complex decision logic that requires careful monitoring and evaluation.

**Benefits**: Provides sophisticated reasoning capabilities for complex scenarios, maintains high early-phase performance while ensuring eventual fairness, and adapts to changing project conditions automatically.

**Use case**: General-purpose applications requiring both performance and fairness, long-term projects where relationship maintenance matters, and scenarios where both efficiency and equity have importance.

**3) Aggressive Balance Mode: Equity-First Philosophy**

**Philosophy**: Prioritizes fair workload distribution and equal robot utilization, operating under the principle that long-term system sustainability requires equitable resource allocation.

**Priority weighting**: Workload fairness (70%) vs. Task success (30%): The 70:30 split creates strong equity bias while maintaining acceptable performance standards. This weighting ensures effective load balancing even when it requires accepting moderate performance reductions.

**Trade-off approach**: Frequently selects robots with lower success rates if they have significantly fewer current tasks, using workload deviation as the primary allocation criterion and treating success rates as secondary considerations.**Accepts**: Success rate reductions of 10-20% to maintain fair distribution, occasional assignment of tasks to less-optimal robots, and potential short-term performance impacts for long-term equity goals.

**Benefits**: Achieves even wear patterns across all robots, improves overall system capacity utilization, reduces single-point-of-failure risks, and ensures all robots gain operational experience across diverse tasks.

**Use case**: Long-term sustainability scenarios where robot longevity matters, environments where all robots must maintain operational readiness, and applications where preventing robot underutilization is critical for system resilience.

**Trade-off Decision Logic**

The trade-off decision thresholds are anchored to the fundamental workload structure where each robot's target load equals 12 tasks (36 total tasks ÷ 3 robots). The 25% success threshold establishes that when success rate gaps between robots exceed 25%, performance differences are so significant that they justify workload imbalances. This threshold derives from the principle that a 25% success rate is equivalent to a 3-task imbalance in project impact (12 × 0.25), representing a substantial operational difference that warrants prioritizing performance over balance.

The 5% lower threshold defines the point where success rate differences become negligible (e.g., Light Robot 65% vs Medium Robot 60% = 5% gap), making workload balance the determining factor since performance differentials are minimal.

**Mode-specific intermediate thresholds** (Success-Focused: 8%, Aggressive Balance: 18%) partition the 5%–25% range according to each mode's strategic philosophy:

- **8% threshold**: Success-Focused mode chooses higher success for gaps >8%, reflecting its first performance approach

- **18% threshold**: Aggressive Balance mode chooses fewer tasks for gaps <18%, demonstrating its equity-first philosophy

Mode-specific intermediate thresholds reflect strategic philosophy: Success-Focused mode's 8% threshold ( 1 task impact) represents the minimum performance gain justifying workload imbalance, while Aggressive Balance mode's 18% threshold ( 2 task impact) represents the maximum performance sacrifice acceptable to maintain fairness.

This graduated threshold system ensures that clear-cut scenarios (>25% or <5% gaps) receive universal treatment, while intermediate cases (5%–25% gaps) are resolved according to the selected strategic mode, providing systematic decision-making across all possible success rate differentials.

**Computational Parameters**

This methodology provides a comprehensive framework for human-like reasoning in multi-robot task allocation while maintaining quantitative performance evaluation capabilities comparable to traditional optimization approaches.

Key system parameters include:

1. **the Maximum Retries**: three per task allocation - this parameter directly corresponds to the three-tier hierarchical retry system described in Section Hierarchical Retry Escalation Strategy. Each retry attempt provides progressively more detailed guidance (Tier 1: specific feedback, Tier 2: comprehensive guidance with examples, Tier 3: maximum support with structured templates) before implementing fallback allocation. The three-attempt limit ensures systematic quality improvement while preventing excessive computational overhead and prolonged cycles.

2. **LLM used: Claude-4-sonnet and LLM Temperature: 0.1 for deterministic reasoning** - This low temperature setting minimizes randomness in LLM outputs, ensuring consistent and reproducible allocation decisions. Since task allocation requires reliable decision-making rather than creative generation, the low temperature promotes deterministic reasoning patterns that align with the structured decision framework and validation requirements of the system.

3. **Recursion Limit: 15 × number of tasks** - This limit accommodates the maximum possible workflow iterations, accounting for the worst-case scenario where each task requires multiple retry attempts. With 36 tasks and up to 3 retries per task, the limit provides substantial computational headroom (15 × 36 = 540 iterations) while preventing infinite loops in case of systematic LLM failures or workflow errors.

4. **Quality Threshold: 0.6-0.8 for validation acceptance** - This range establishes the minimum acceptable quality score for LLM reasoning validation as described in node 4. Scores below 0.6 indicate significant deficiencies requiring retry intervention, while scores above 0.8 represent high-quality reasoning. The range allows for configurable strictness depending on application requirements, balancing quality assurance with computational efficiency.

**Multi-Robot Allocation State Persistence**

The framework employs LangGraph's Memory Saver functionality maintain state persistence throughout the allocation process, ensuring that complex multi-robot allocation workflows can survive interruptions such as LLM API failures, timeout errors, or system crashes without losing critical decision context. The persistent storage structure includes current task queue position, robot workload distributions, allocation history with reasoning explanations, balance scores and phase progress, validation results and retry attempts, and LLM interaction records. This comprehensive state preservation is essential for multi-robot allocation because workload balance requires that early allocation phase decisions incrementally across the entire task sequence.

When workflow interruptions occur, for instance, consider the system processing task 18 of 36

tasks in the middle allocation phase with Light Robot assigned 10 tasks, Medium Robot assigned 5 tasks, Heavy Robot assigned 3 tasks, balance score of 75, and phase-adaptive weights favoring 60% success and 40% workload balance. Without state persistence, system restart would force allocation to begin from task 1 with all robots at zero tasks, potentially leading to inconsistent decisions that ignore previously established patterns. The state persistence mechanism enables seamless resumption from task 18 with all accumulated information intact.

The recovery mechanism validates state integrity by confirming task queue accuracy, verifying robot workload calculations, and ensuring phase detection consistency before resuming. The procedure includes safeguards for preventing common restoration errors such as double counting completed tasks or resetting phase progression inappropriately.

## RESULTS

The evaluation of the LTAA framework proceeded through systematic validation establishing: (1) LLM feasibility for multi-robot coordination through progressive architecture development, (2) computational efficiency gains via dynamic prompting and focused scope, and (3) competitive performance with traditional optimization algorithms. This section presents results in three parts corresponding to the framework development trajectory: feasibility assessment establishing foundational viability, computational efficiency analysis demonstrating architectural improvements, and performance benchmarking validating competitive allocation effectiveness.

**SMART-LLM Feasibility Assessment Results**

The SMART-LLM implementation study revealed fundamental challenges in LLM-driven coordination that motivated subsequent architectural development. Table 1 presented in Section Framework Design Rationale documents baseline performance across four task complexity categories using the 36 SMART-LLM benchmark tasks.

Execution Failures are indicated by The asterisk notation in Table 1 quantifies complete execution failures where tasks could not run at all due to fundamental errors. Elemental tasks experienced 2 execution failures, Simple tasks showed 5 failures, while both Compound and Complex categories each encountered 6 complete failures. These execution breakdowns stemmed from error sources documented in Appendix 1 Table 9

Overall Performance of The framework achieved success rates ranging from 0.10 for Compound tasks to 0.56 for Elemental tasks, demonstrating performance degradation as coordination complexity increased. Executability metrics ranged from 0.39 to 0.71 across categories, indicating that even when tasks executed, many failed to achieve valid allocation outcomes. Robot Utilization scores between 0.56 and 0.66 revealed suboptimal coordination efficiency across all complexity levels.

**Self Corrective-Agent Architecture Results**

The Self Corrective-Agent Architecture evaluation demonstrated that systematic validation and hierarchical retry mechanisms could address the reliability challenges identified during SMART-LLM replication. Table 2 presented in Section Performance Validation and Evaluation Limitations documents performance metrics across the same 36-task SMART-LLM benchmark under identical evaluation conditions, isolating the impact of architectural improvements.

The most significant finding is the complete elimination of execution failures all 36 tasks executed successfully without the errors that plagued SMART-LLM implementation. The absence of asterisks in Table 2 compared to Table 1's numerous execution failures validates the effectiveness of multi-stage validation and structured feedback loops. This 100% execution rate demonstrates that systematic validation can transform unreliable LLM outputs into consistent, executable coordination plans.

The framework achieved success rates ranging from 0.21 (Compound tasks) to 0.50 (Elemental and Complex tasks). More significantly, Executability improved substantially across all categories: Elemental (0.75), Simple (0.83), Compound (0.88), and Complex (0.86). These executability gains of 9-30 percentage points demonstrate that validation mechanisms ensured syntactically and semantically correct outputs.

The Self Corrective-Agent Architecture demonstrates computational efficiency gains despite using the same few-shot prompting strategy as SMART-LLM. The framework consumes 700,017

53　　　　　　　　　　　　　　　kaitha, December 1, 2025

total tokens versus SMART-LLM's 727,649 tokens (3.8% reduction) as shown in the figure 13 and completes execution in 746 seconds versus 1,075 seconds (30.6% reduction) shown in the figure 14. This improvement is primarily due to its modular prompting design, where each stage (decomposition, allocation, or code generation) loads only its corresponding example file. In contrast, SMART-LLM concatenates all three example files into every LLM call, regardless of which stage is active. As a result, each Self Corrective-Agent inference processes far fewer tokens, enabling significantly faster reasoning. Notably, this stage-specific prompting structure is so efficient that even with validation-driven retries, Self Corrective-Agent still consumes fewer tokens overall.

Additionally, SMART-LLM lacks internal validation mechanisms, so when an error occurs at any step, the entire pipeline must be rerun from the beginning, repeatedly incurring the computational cost of its large concatenated files. This contrasts with Self corrective-Agent's localized error handling, where only the failing stage is regenerated while prior outputs are preserved. Thus, the absence of validation in SMART-LLM not only reduces reliability but also leads to higher token consumption due to repeated full-task reruns.

Beyond the token savings achieved through modular design the Self Corrective-Agent Architecture demonstrates 30.6% execution time improvement over SMART-LLM (746 vs. 1,075 seconds). This reduces tokens per call (one example file vs. three concatenated files), decreasing LLM inference time and API processing overhead per invocation. Combined with the 3.8% token reduction from targeted retries, the modular structure achieves both improved reliability and faster execution time compared to SMART-LLM's monolithic prompting approach.

**LTAA Framework Evaluation: Computational Efficiency and Performance Validation**

*Computational Efficiency : Token Consumption*

The framework evolution demonstrates substantial improvements in computational efficiency through architectural refinement. Figure 13 presents token consumption across three framework iterations, while Figure 14 shows corresponding execution time metrics.

As illustrated in Figure 13, LTAA achieves token reduction compared to full-pipeline frame-works. LTAA consumes 39,188 total tokens across 36 tasks (1,088 tokens per task), representing a 94.6% reduction from SMART-LLM (727,649 tokens, 20,212 per task) and 94.4% reduction from Self Corrective-Agent Architecture (700,017 tokens, 19,445 per task).

This substantial improvement stems from complementary architectural innovations rather than dataset characteristics. To enable fair comparison, we analyze SMART-LLM's token consumption by computational stage. SMART-LLM processes tasks through three sequential stages task decomposition, allocation, and code generation consuming 727,649 total tokens across 36 tasks (20,212 tokens per task). Assuming approximately equal token distribution across stages, each stage consumes roughly 242,883 tokens total, or 6,747 tokens per task for allocation alone. Even when comparing LTAA's allocation-only scope (1,088 tokens per task) against this estimated allocation component of SMART-LLM (6,747 tokens per task), LTAA achieves 84% token reduction.

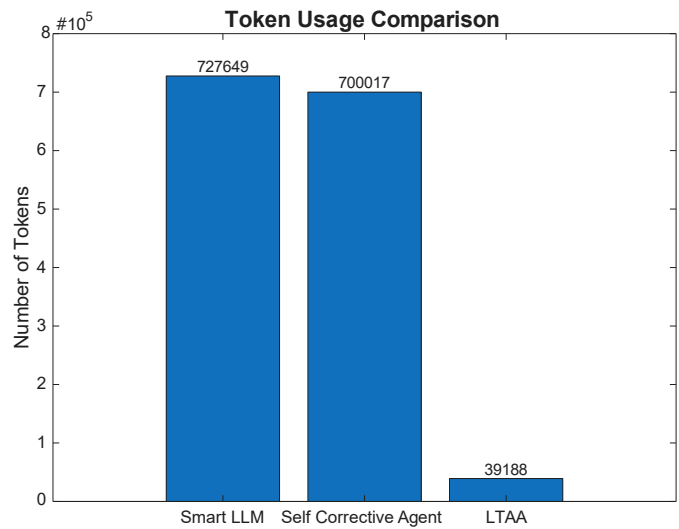

**Fig. 13.** Token Consumption Comparison.

This efficiency gain stems primarily from dynamic prompt generation: LTAA constructs task-specific prompts incorporating only current system state (task features, robot capabilities, workload context, phase guidance), eliminating the 5,000-6,000 tokens of redundant few-shot examples concatenated into every SMART-LLM prompt regardless of task requirements.

The token reduction reflects two complementary innovations: (1) focused allocation scope eliminating decomposition and code generation overhead ( 14,000 tokens per task), and (2) dynamic prompt generation replacing few-shot example concatenation with structured context provision, reducing allocation-stage prompts from 6,700 to 1,100 tokens.

*Computational Efficiency: Task Allocation Time*

Figure 14 demonstrates corresponding improvements in execution time. LTAA completes 36 task allocations in 149 seconds (4.14 seconds per task), achieving 86.1% time reduction compared to SMART-LLM (1,075 seconds, 30.00 seconds per task) and 80.0% reduction compared to Self Corrective-Agent Architecture (746 seconds, 21.00 seconds per task).

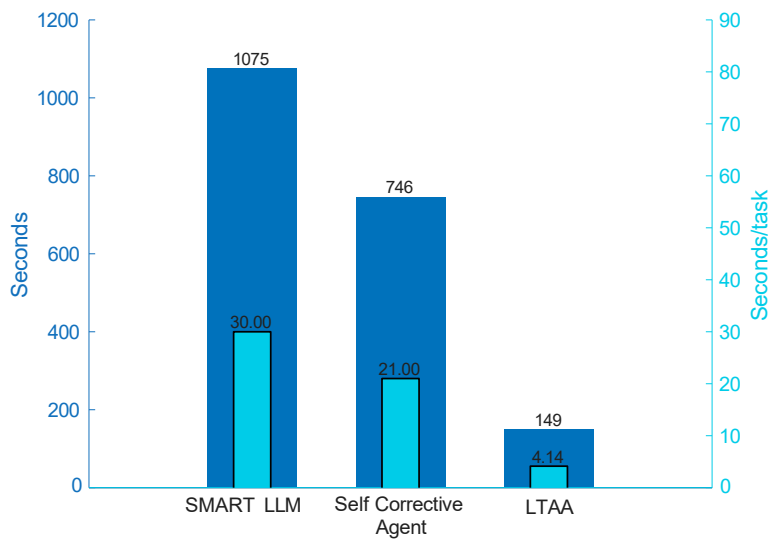

**Fig. 14.** Task Allocation Time Comparison by Methods.

To enable comparison at equivalent functional scope, we isolate SMART-LLM's allocation-stage execution time. SMART-LLM executes three sequential stages consuming 30 seconds per task total. Assuming approximately equal time distribution across stages yields roughly 10 seconds per task for the allocation stage alone (1,075 total seconds ÷ 3 stages ÷ 36 tasks). Comparing allocation stages directly LTAA (4.14 seconds per task) versus SMART-LLM allocation component (10 seconds per task) reveals 59% execution time reduction even at equivalent scope.

This stage-level efficiency stems from reduced token processing LTAA's dynamic prompts (1,088 tokens) process faster than SMART-LLM's concatenated three-file prompts ( 6,747 tokens per stage), directly reducing LLM inference time and API communication latency, and elimination of redundant example file loading LTAA constructs task-specific prompts without loading concatenated example files, while SMART-LLM loads all three example files (decomposition, allocation, code generation) for every call regardless of stage relevance. Combined with focused scope eliminating decomposition and code generation stages entirely, LTAA achieves 86.1% overall execution time improvement through architectural innovations rather than reduced functional capability.

**LTAA Framework Performance Benchmarking**

The evaluation of the LLM-based task allocation framework demonstrates its effectiveness in multi-robot allocation scenarios. Several key metrics, including overall success rates, workload distribution, and adaptability across different operational contexts, were assessed. The framework's performance was evaluated through comprehensive testing across traditional optimization methods, multi-scenario adaptability, operational mode effectiveness, and feature-specific allocation patterns.

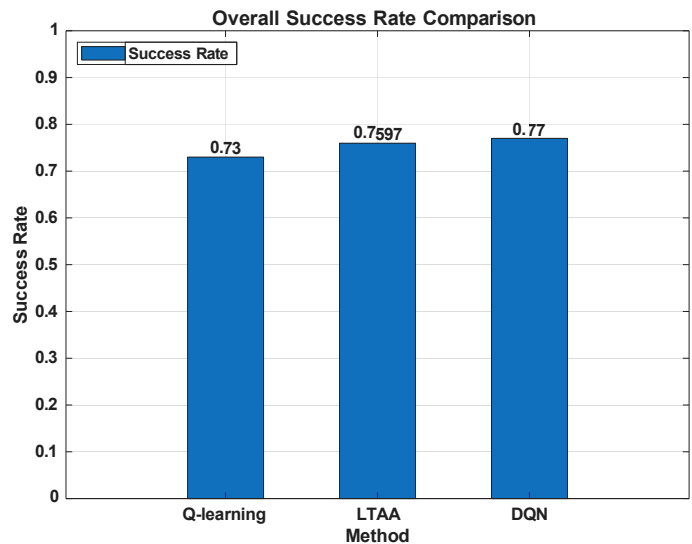

**Fig. 15.** Task Allocation Methods Comparison.

*Task Allocation Methods Comparison*

From the figure 15 the LLM-based framework achieves competitive performance with an overall success rate of 75.97%, positioning it between Q-learning (73%) and DQN (77%) optimization approaches. This demonstrates that reasoning-based allocation can achieve results comparable to traditional algorithmic methods while providing enhanced interpretability and adaptability. The 2.97% improvement over Q-learning and the narrow 1.03% gap with DQN indicates the viability of LLM integration for complex allocation. However, DQN reaches this performance with extensive model parameter tuning, while LTAA demonstrated high zero-shot transferrability with minimal tuning efforts needed.

*Multi-Scenario Performance Analysis*

To evaluate framework robustness across varying robot team compositions, three scenarios with distinct capability distributions were tested. Each scenario emphasizes different robot specializations by adjusting success rate matrices while maintaining the same 36-task allocation sequence.

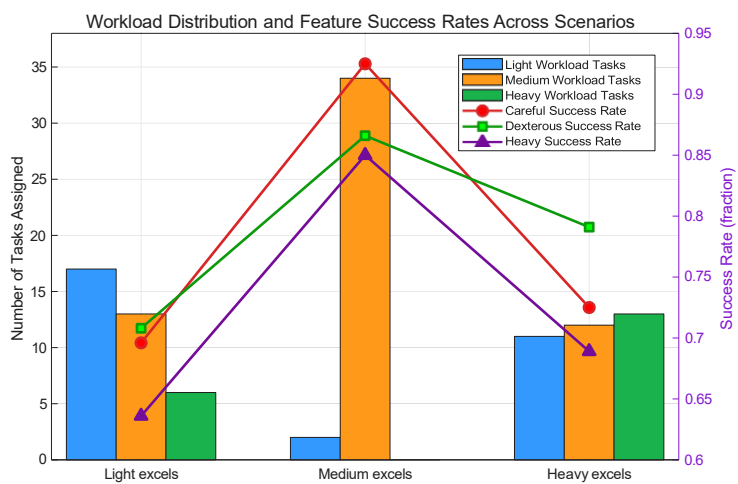

**Fig. 16.** Multi-Scenario Performance Analysis .

As shown in Figure 16, the Heavy Excels scenario demonstrates the framework's ability to achieve both strong performance and workload equity. With 77.1% overall success rate and balanced workload distribution (11, 12, 13), this configuration validates that the framework can maintain competitive performance while ensuring fair task allocation across all robots a critical requirement for construction environments where both efficiency and equitable resource utilization matter.

The Medium Excels scenario achieves the highest overall success rate (87.8%), leveraging the balanced capabilities of medium robots with task distribution (2, 34, 0). The extreme workload concentration occurs because the LLM consistently selects the balanced-generalist robot when it demonstrates competitive success rates across all task types. Feature-specific success rates in this scenario are: careful tasks (92.5%), dexterous tasks (86.6%), and heavy tasks (85%). While this demonstrates maximum performance optimization, the workload imbalance illustrates the performance-fairness trade-off inherent in allocation decisions.

The Light Excels scenario, while showing lower overall success (69.9%), maintains reasonable workload balance (17, 13, 6) and consistent feature performance across task types, demonstrating that the framework adapts its allocation strategy even when the specialized robot has moderate rather than dominant capabilities.

*Operational Mode Performance Analysis*

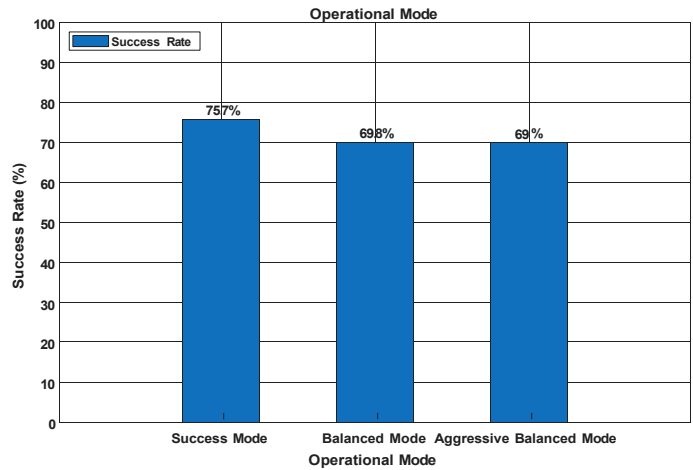

**Fig. 17.** Operational Mode Performance Comparison .

As illustrated from the figure 17 the three operational modes demonstrate distinct performance characteristics aligned with their strategic philosophies. Success-Focused Mode achieves the highest success rate (75.7%), confirming its performance-first approach. Balanced Mode (69.8%) and Aggressive Balance Mode (69%) show similar success rates, indicating that equity-focused strategies maintain acceptable performance levels while prioritizing fairness objectives. The moderate performance trade-off (approximately 6% reduction) demonstrates the framework's ability to balance competing objectives effectively.

*Workload Distribution and Feature Success Analysis*

As shown in the figure 18 the workload distribution reveals strategic allocation patterns with Light Robot handling the majority of tasks (22), followed by Medium Robot (10) and Heavy Robot (4). This distribution reflects the framework's optimization for robot capability matching.

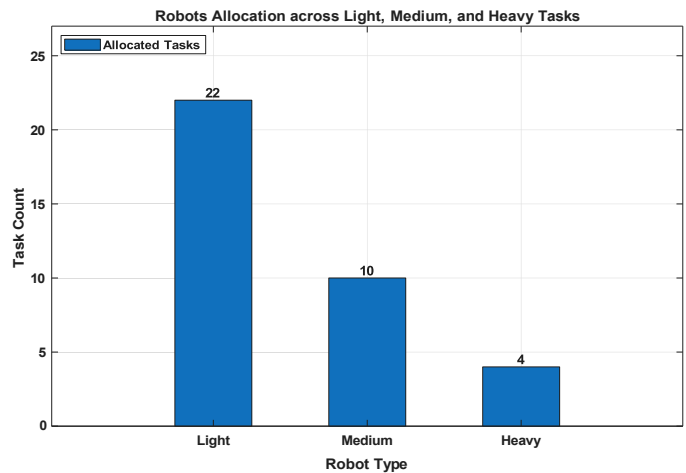

**Fig. 18.** Workload Distribution by Robot Type.

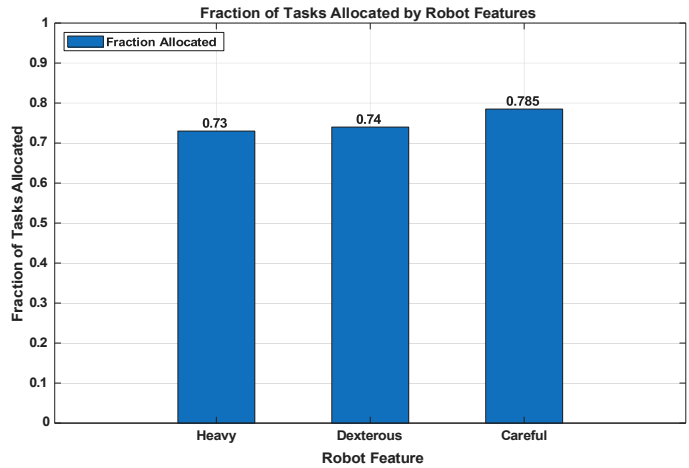

**Fig. 19.** Workload Distribution and Feature Success comparison.

From the figure 19, Feature-specific success rates show strong performance across all categories: Heavy tasks (73%), Dexterous tasks (74%), and Careful tasks (78.5%). The high success rate for careful tasks aligns with the prevalence of precision-oriented allocations, while the framework maintains robust performance across diverse task requirements.

Brute force, greedy, and DP obtained success rates of 0.77, 0.81, and 0.95. However, they are omitted from the comparison plots because these deterministic algorithms lack uncertainty modeling. Given that LTAA, Q-learning, and DQN operate under stochastic conditions, RL approaches provide the appropriate basis for direct comparison.

## CONCLUSIONS

This study addressed a critical gap in multi-robot task allocation by developing the LTAA framework through a systematic progression of feasibility validation, reliability enhancement, and algorithmic benchmarking. The initial SMART-LLM replication demonstrated that plain LLM prompting suffers from inconsistency, execution failures, and simulation incompatibilities, highlighting the need for structured validation. The subsequent Self Corrective-Agent Architecture resolved these limitations by introducing multi-stage validation and localized retry mechanisms, eliminating execution failures entirely and improving computational efficiency through modular prompting. This stage-specific prompting reduced token usage and reasoning time even when retries were required.

Building on these foundations, the LTAA framework advanced both computational performance and allocation quality. Dynamic prompt generation reduced token consumption by 94.6% and cut average allocation time by 86%, demonstrating substantial efficiency gains over SMART-LLM. Benchmarking against traditional algorithms further confirmed LTAA's effectiveness: it achieved a 75.97% overall success rate without any training or fine-tuning, performing competitively between Q-learning (73%) and DQN (77%). Scenario-level analysis showed particularly strong performance in the Heavy Excels setting (77.1% success with balanced workloads), while operational mode testing validated its ability to manage performance fairness trade offs.

Compared to approach with SMART-LLM, LTAA is significantly more robust it exhibits zero execution failures, faster reasoning, and substantially lower computational overhead. These combined results demonstrate that LTAA not only overcomes the reliability and scaling limitations of SMART-LLM but also provides a practical, interpretable, and computationally efficient alternative to traditional optimization-based MRTA methods.

## DISCUSSION

This research establishes several key novelties for construction robotics task allocation. First, it proposes a phase-adaptive allocation strategy dynamically shifts priorities throughout project progression. Second, it included more comprehensive workload allocation goals, transitioning from execution feasibility-only goals to workload and robot usage balances. Moreover, the multi-stage validation framework with hierarchical retry mechanisms eliminates execution failures. It transforms unreliable LLM outputs into consistent coordination plans through systematic quality assurance. Most significantly, systematic benchmarking demonstrates competitive performance with traditional algorithms. LLM-based reasoning provides natural language interpretability and rapid adaptability without retraining. This challenges the assumed trade-off between transparency and effectiveness in multi-robot coordination.

However, this study has several important limitations. Firstly, the feasibility assessment phase employed a different dataset (AI2-THOR) than the benchmarking phase (Tasks from TEACh dataset). While this transition was methodologically necessary enabling direct replication for feasibility assessment and providing common ground for fair algorithmic comparison, it provided an indirect performance comparison with tasks with similar nature. However, the benchmark of LTAA agaist traditional MRTA approaches used the same dataset. The authors are also working on testing the SMART-LLM in TEACh dataset for a more balanced benchmark system.

Secondly, this study adopted a relatively small evaluation scale. While this enabled systematic baseline comparison and thorough capability analysis, real-world construction involves substantially more tasks with complex spatial and temporal constraints not fully captured in the evaluation dataset. This study used some abstract features such as elemental/compound, heavy/dexterous to represent the construction tasks, instead of working on specific tasks for better representation and generalization. The abstract was also necessary considering the lack of a large-scale real-world construction operation database. The authors believe with such a database, evaluating the LTAA in additional scenarios involving dynamic replanning and resource conflicts would strengthen confidence in framework robustness.

Moreover, LLM output variability presents reliability concerns for safety-critical construction operations. While multi-stage validation eliminated execution failures, controlled stochasticity enables nuanced reasoning but potentially produces allocation variations. Construction environments requiring absolute determinism may need additional verification mechanisms beyond current validation. Furthermore, when comparing the computational resources needed, SMART-LLM's token and time costs could not be measured per stage, because all three stages share a single concatenated prompt. To enable comparison with LTAA, which evaluates only allocation, we approximated SMART-LLM's stage-level cost by dividing its total tokens and runtime by three. This approximation is reasonable but may not fully capture the true allocation-specific overhead, and future work with stage-level instrumentation would allow more precise analysis.

Future research priorities should focus on conducting comprehensive trade-off studies to evaluate the framework across extended performance-fairness scenarios, enabling more nuanced under-standing of optimal allocation strategies for different construction contexts. Real-world validation through physical robot demonstrations and construction site implementations will be essential to verify the framework's practical applicability and address reliability concerns. Additionally, developing hybrid approaches that integrate structured optimization frameworks with LLM reasoning capabilities could enhance both computational rigor and contextual adaptability, advancing the field toward more robust and interpretable construction automation systems.

## ACKNOWLEDGMENT

This study has not received external financial support.

## APPENDIX 1. CHALLENGES ENCOUNTERED DURING SMART-LLM STUDY

**TABLE 9.** Implementation Challenges Identified During SMART-LLM Study

| Challenge | Description & Root Cause | Impact on System Performance | LTAA Framework Response |
|---|---|---|---|
| **1. State Validation Inconsistency** | AI2-THOR's geometric threshold validation failed to detect correctly placed objects despite visual confirmation of proper positioning | False negative task completions; underestimated success rates in object placement tasks (drawers, sinks) | Multi-stage validation system (Node 4) with 8-rule framework ensuring output quality verification |
| **2. Object Ambiguity in Multi-Instance Environments** | Detection system failed to disambiguate between identical objects (e.g., bathtub faucet vs. sink faucet) due to lack of spatial-contextual reasoning | Wrong object selection causing complete task failures; agents acted on incorrect similar objects | Context-aware prompt generation (Node 2) incorporating spatial relationships and task intent |
| **3. Lack of Inter-Agent Object Transfer Mechanism** | No coordination protocol for sequential object manipulation across agents; first agent retained object preventing second agent execution | Complete failure in collaborative sequential tasks; tasks requiring object handoff could not be completed | Explicit state management through PipelineState schema (tracking object ownership and transfer status |
| **4. Navigation Failures** | Overly restrictive goal threshold parameter (0.25m) caused agents to oscillate near target without completion in cluttered environments | Repetitive circular motions; prolonged navigation time; timeout errors preventing task progression | This challenge motivated LTAA's allocation-level evaluation using TEACH dataset, enabling direct comparison with optimization algorithms |
| **5. Object Name Mismatch** | Inconsistency between task-level natural language identifiers (e.g., "trashcan") and AI2-THOR internal names (e.g., "GarbageCan"); case-sensitivity issues | Silent lookup failures; agents executed correctly but failed completion criteria due to name mismatches | Semantic normalization layer with name-mapping and synonym resolution mechanisms |
| **6. Dataset Formatting & Integrity Issues** | Malformed JSON structures with missing essential fields (object states, args); syntax errors (missing commas, brackets) | Parse failures preventing dataset loading; 22% of tasks initially unloadable; required manual schema validation | Comprehensive preprocessing pipeline with automated schema validation and integrity checks |

# APPENDIX 2

```
=== SYSTEM PROMPT (sent to LLM) ===
You are SMART-LLM, an intelligent task allocation expert who thinks like an experienced project manager.
Your strengths:

- Contextual understanding of nuanced situations
- Adaptive reasoning that adjusts to new information
- Clear, explainable decision-making
- Sound judgment in ambiguous gray areas

You make allocation decisions by analyzing trade-offs between success rates and workload balance, providing clear reasoning for
every choice. You adapt your strategy based on the current phase and operational mode.
You are NOT a calculator or algorithm. You are an intelligent agent making thoughtful, explainable decisions based on judgment
and reasoning.
```

**Fig. 20.** Example Systems prompt

```
*Features**: [dexterous]

**Robot Analysis**:
- Light Robot: 80% success (current: 0 tasks, deviation: -12)
- Medium Robot: 60% success (current: 0 tasks, deviation: -12)
- Heavy Robot: 40% success (current: 0 tasks, deviation: -12)

**Decision**: Light Robot

**Expected Success**: 80%

**Reasoning**:
This is a clear-cut decision in SUCCESS_FOCUSED mode. The Light Robot has a significant 20% advantage
over Medium (80% vs 60%) and a massive 40% advantage over Heavy Robot. Since we're in the EARLY
phase with all robots at zero tasks, there's no workload imbalance to consider. The Light Robot's precision
specialization makes it the natural choice for dexterous tasks like ""Stop"" operations that likely require
careful, controlled movements.

**Workload After Assignment**:
- Light: 1 tasks
- Medium: 0 tasks
- Heavy: 0 tasks

**Confidence Level**: High

**Trade-Off Summary**: No meaningful trade-offs required - optimal robot for the task with no workload
constraints to consider.",natural_language,0,True
1,Move to,Motion,['dexterous'],light_robot,0.8,"Despite Light Robot already having 1 task, the 20% success
advantage (80% vs 60%) is significant and aligns with our SUCCESS_FOCUSED mode priority. In the early
phase, establishing a strong foundation of successful task completion outweighs workload balance concerns.
The dexterous requirement clearly favors Light Robot's precision capabilities, and while Medium/Heavy
robots are underutilized, the success gap is too large to justify sacrificing performance for balance at this
stage.",
High,"Accepted continued workload imbalance to secure 20% higher success rate, prioritizing mission success
over equal distribution in early phase",1.0,"## Task 1: Move to
```

**Fig. 21.** Example output of one task of how the reasoning is made

The following three figures (Figures 22–24) present a complete example of the dynamically generated user prompt for a single task allocation decision. The prompt is displayed across three figures due to its comprehensive nature and page layout constraints. Figure 22 shows the task specification and robot capability assessment component, Figure 23 presents the workload context and phase-specific guidance, and Figure 24 illustrates the decision framework and output template. Together, these demonstrate how the system integrates task specifications, robot capabilities, workload state, and phase-specific guidance into a unified reasoning context for the LLM-based allocation agent.

```
=== USER PROMPT (sent to LLM) ===
You are a task allocation expert who uses judgment and reasoning, not formulas.

Your job is simple: Assign this task to a robot in a way that maximizes success while keeping workload fair. Think through the
trade-offs like an experienced project manager would.

## CURRENT SITUATION

**Task to Allocate**: #0 "Stop"
**Features Required**: [dexterous]

**Current Workload Status**:
- Light Robot: 0 tasks (target: 12, deviation: -12)
- Medium Robot: 0 tasks (target: 12, deviation: -12)
- Heavy Robot: 0 tasks (target: 12, deviation: -12)

**Allocation Progress**: EARLY phase (36 tasks remaining)
**Mode**: BALANCED

## ROBOT CAPABILITIES & SUCCESS RATES FOR THIS TASK

**CALCULATED SUCCESS RATES**
- **Light Robot**: 80% success for this task
- **Medium Robot**: 60% success for this task
- **Heavy Robot**: 40% success for this task

**General Robot Capabilities** (for context):
- **Light Robot** (Precision Specialist): Strong at dexterous (80%) and careful (90%) tasks, weak at heavy (30%)
- **Medium Robot** (Balanced Generalist): Decent at all tasks - heavy (70%), dexterous (60%), careful (70%)
- **Heavy Robot** (Force Specialist): Strong at heavy (90%) tasks, weak at dexterous (40%) and careful (50%)
```

**Fig. 22.** User prompt example: Task specification and robot capability assessment (Part 1 of 3)

## YOUR DECISION FRAMEWORK

**BALANCED Mode Behavior** (Recommended):
Your goal is high success with reasonable workload fairness.
- Balance when success gap is small-moderate (<8%)
- Accept 7-10% success drops if imbalance is severe
- Think sequentially about future task distribution
- Aim for ±4 tasks from target per robot

Example reasoning: "Light (82% success, 14 tasks) vs Medium (75% success, 8 tasks). That's 7% success gap and 6 task imbalance. Balance gain justifies modest success sacrifice."

**EARLY PHASE Strategy** (First 30% of tasks):
**Mindset**: Build a strong foundation
- Prioritize success rate more heavily
- Accept some imbalance initially
- Focus on avoiding terrible mismatches

Typical reasoning: "Early phase: Establishing strong baseline success rate. Will address balance in middle/late phases once foundation is solid."

### Step 1: Understand the Task
What does this task require? Look at the features:
- **Heavy**: Strength, force, lifting capacity
- **Dexterous**: Precision, fine motor skills, careful manipulation
- **Careful**: Gentle handling, fragility awareness, damage prevention
### Step 2: Evaluate Robot Suitability
The success rates are already calculated (see above).
For each robot, consider:
- "Is this success rate acceptable for the task criticality?"
- "Are there any severe mismatches?" (Red flags: <50% success)
- "What's the success gap between robots?"
### Step 3: Check Workload Balance
Look at the current distribution:
- Which robots are overloaded? (>5 tasks above target)
- Which robots are underutilized? (>3 tasks below target)
- How severe is the imbalance?
### Step 4: Make the Trade-Off Decision
This is where your judgment matters most:

**If success gap >20%**: Always prioritize success. Mention imbalance but proceed.
**If success gap 10-20%**: Weigh the trade-off. Consider criticality, imbalance severity, phase, future tasks.
**If success gap <10%**: Favor the underutilized robot. Success difference is small.
**If success gap <5%**: Always choose robot with fewer tasks.

### Step 5: Look Ahead (Sequential Thinking)
Before finalizing, consider:
- "If I assign this to Robot A, what tasks are left for it?"
- "Will Robot A get overloaded with similar future tasks?"
- "Would saving Robot A's capacity be smarter?"

**Fig. 23.** User prompt example: Workload context and phase guidance (Part 2 of 3)

## REQUIRED OUTPUT FORMAT

Respond in this exact markdown format:

```markdown
## Task 0: Stop
**Features**: [dexterous]

**Robot Analysis**:
- Light Robot: 80% success (current: 0 tasks, deviation: -12)
- Medium Robot: 60% success (current: 0 tasks, deviation: -12)
- Heavy Robot: 40% success (current: 0 tasks, deviation: -12)

**Decision**: [Chosen Robot]
**Expected Success**: [X%]

**Reasoning**:
[2-4 sentences explaining why this robot was chosen, what trade-offs were made if any, and how this fits into the overall allocation strategy. Be specific about success gaps and workload considerations.]

**Workload After Assignment**:
- Light: [X] tasks
- Medium: [Y] tasks
- Heavy: [Z] tasks

**Confidence Level**: [High/Medium/Low]

**Trade-Off Summary**: [What was sacrificed, if anything, and why it was acceptable]
```
## CRITICAL REMINDERS
✅ **DO**:
- Think like a project manager, not a calculator
- Explain every trade-off you make
- Consider full context (phase, mode, remaining tasks)
- Use judgment for gray areas (8-15% success gaps)
- Look ahead at future task implications
- State confidence levels honestly

❌ **DON'T**:
- Calculate exact cost values or use formulas
- Make decisions without explaining why
- Ignore severe workload imbalances
- Sacrifice >20% success for balance (except aggressive mode)
- Assign tasks that create safety risks

Remember: You're allocating robot tasks, not solving math problems. Trust your reasoning and explain it clearly!

Total prompt length: 5724 characters

**Fig. 24.** User prompt example: Decision framework and output template (Part 3 of 3)

using an improved genetic algorithm. *Automation in Construction*, 165, 105583.

[29] Gerkey, B. P., & Matarić, M. J. (2004). A formal analysis and taxonomy of task allocation in multi-robot systems. *The International Journal of Robotics Research*, 23(9), 939-954.

[30] Korsah, G. A., Stentz, A., & Dias, M. B. (2013). A comprehensive taxonomy for multi-robot task allocation. *The International Journal of Robotics Research*, 32(12), 1495-1512.

[31] Kuhn, H. W. (1955). The Hungarian method for the assignment problem. *Naval Research Logistics Quarterly*, 2(1-2), 83-97.

[32] Atay, N. (2006). Mixed-integer linear programming solution to multi-robot task allocation problem. *Technical Report*, Carnegie Mellon University.

[33] Jones, E. G. (2020). Multi-agent coordination for disaster response with intra-path precedence constraints based on genetic algorithm. *Computers & Operations Research*, 124, 105075.

[34] Chen, J. (2019). Applied ant colony optimization to cooperative task allocation of heterogeneous unmanned aerial vehicles. *Swarm and Evolutionary Computation*, 47, 168-187.

[35] Feng, C., Xiao, Y., Willette, A., McGee, W., & Kamat, V. R. (2016). Vision guided autonomous robotic assembly and as-built scanning on unstructured construction sites. *Automation in Construction*, 59, 128-138.

[36] Liang, C. J., & Cheng, T. (2023). Adaptive construction robot control in unstructured environments. *Journal of Construction Engineering and Management*, 149(3), 04022168.

[37] Prieto, S. A., & García de Soto, B. (2024). *Collaborative Large Language Models for Task Allocation in Construction Robots*. Available at SSRN 5097309.

[38] Chen, J., Yu, C., Zhou, X., Xu, T., Mu, Y., Hu, M., Shao, W., Wang, Y., Li, G., & Shao, L. (2024). *EMoS: Embodiment-aware heterogeneous multi-robot operating system with LLM agents*. arXiv:2410.22662.

[39] Deng, M., Fu, B., Li, L., & Wang, X. (2025). *Integrating LLMs and Digital Twins for Adaptive Multi-Robot Task Allocation in Construction*. arXiv:2506.18178.

[40] Makondo, N., Choudhury, S., & Sridharan, M. (2015). Integrated task and motion planning for mobile manipulation. *Robotics and Autonomous Systems*, 74, 119-131.

72    kaitha, December 1, 2025